\documentclass{ieeeaccess_moded}
\DeclareUnicodeCharacter{2217}{\ensuremath{\ast}}
\usepackage{amsthm,url}
\newtheorem{definition}{Definition}
\newtheorem{theorem}{Theorem}

\usepackage{physics}
\usepackage{caption}
\usepackage{subcaption}
\usepackage{amsfonts}
\usepackage{xcolor}
\usepackage{multirow}
\usepackage{wrapfig}
\usepackage{array}
\usepackage{placeins}
\usepackage{balance}

\usepackage{pifont}

\usepackage{cite}
\usepackage{epstopdf}
\usepackage{amsmath,amssymb,amsfonts}
\usepackage{algorithms}
\usepackage{graphicx}
\usepackage{textcomp}
\usepackage{enumitem}
\def\BibTeX{{\rm B\kern-.05em{\sc i\kern-.025em b}\kern-.08em
    T\kern-.1667em\lower.7ex\hbox{E}\kern-.125emX}}

\usepackage[ruled,linesnumbered]{algorithm2e}
\usepackage{hyperref}

\begin{document}
\title{
Computable Model-Independent Bounds for Adversarial Quantum Machine Learning
}
\author{\uppercase{Bacui Li,}\authorrefmark{1,3}
\uppercase{Tansu Alpcan,\authorrefmark{1} Chandra Thapa,\authorrefmark{3}} and \uppercase{Udaya Parampalli\authorrefmark{2}}. }
\address[1]{Department of Electrical and Electronic Engineering, University of Melbourne, Parkville, Victoria 3010, Australia}
\address[2]{School of Computing and Information Systems, University of Melbourne, Parkville, Victoria 3010, Australia}
\address[3]{CSIRO Data61, Sydney, Australia}


\corresp{Corresponding author: Bacui Li (email: bacuil@student.unimelb.edu.au).}

\begin{abstract}
By leveraging the principles of quantum mechanics, QML opens doors to novel approaches in machine learning and offers potential speedup. However, machine learning models are well-documented to be vulnerable to malicious manipulations, and this susceptibility extends to the models of QML. This situation necessitates a thorough understanding of QML's resilience against adversarial attacks, particularly in an era where quantum computing capabilities are expanding. 
In this regard, this paper examines model-independent bounds on adversarial performance for QML. To the best of our knowledge, we introduce the first computation of an approximate lower bound for adversarial error when evaluating model resilience against sophisticated quantum-based adversarial attacks. Experimental results are compared to the computed bound, demonstrating the potential of QML models to achieve high robustness. In the best case, the experimental error is only 10\% above the estimated bound, offering evidence of the inherent robustness of quantum models. 
This work not only advances our theoretical understanding of quantum model resilience but also provides a precise reference bound for the future development of robust QML algorithms. 
\end{abstract}


\titlepgskip=-15pt
\maketitle

\section{Introduction}
Machine learning (ML) is increasingly integral in various applications such as speech recognition \cite{Graves2013SpeechNetworks}, computer vision \cite{Kendall2017WhatVision}, and natural language processing \cite{Wolf2020Transformers:Processing}. 
A significant concern in ML is its vulnerability to adversarial attacks, in which malicious actors exploit weaknesses in the models to produce outcomes favorable to the adversary~\cite{Carlini2016TowardsNetworks, Goodfellow2014ExplainingExamples, Kurakin2016AdversarialScale}. Moreover, it is demonstrated that various ML classification models are at risk when adversaries craft examples specifically designed to mislead them~\cite{Bruna2013IntriguingNetworks, Biggio2013EvasionTime}.

Alongside classical machine learning (CML), quantum machine learning (QML) has emerged as a promising new paradigm. Conceptually similar to classical counterparts, QML involves iteratively optimizing models across training samples to achieve intended functionalities \cite{Melnikov2023QuantumEngineering}. 
The key distinction between quantum and classical models lies in the use of quantum computing.
QML exploits quantum mechanical phenomena such as superposition and entanglement, which are core principles of quantum computing. The potential of quantum computing has been demonstrated in various domains~\cite{Shor1994AlgorithmsFactoring, Farrell2023ScalableQubits, Santagati2023DrugComputers, Ladd2010QuantumComputers}. Moreover, in the context of ML, quantum systems offer the possibility of manipulating exponentially large state spaces efficiently~\cite{Li2022AClassification}. 

QML shows promise but faces challenges in practical implementation, particularly regarding vulnerability to adversarial attacks. While some studies suggest QML models may exhibit resilience to certain adversarial transfer attacks designed for classical models~\cite{West2023BenchmarkingScale}, the overall security landscape remains open. Studies have primarily focused on adversarial examples using classical input data in trained quantum classification models, observing similar misclassification phenomena to those in classical models \cite{Lu2020QuantumLearning}. However, these insights provide limited guidance for designing robust QML models. The field is further constrained by limitations in classical simulation capabilities and noisy quantum hardware, hindering extensive experimental studies. Theoretically, the security of QML in handling exponentially larger spaces remains debated~\cite{Liu2020VulnerabilityPerturbations}. However, some research indicates that, under certain assumptions, the scaling of security risks in QML is comparable to that in CML~\cite{Liao2021RobustLearning}.

Along with these results, in the context of adversarial attacks, it is crucial to have a fundamental understanding of adversarial performance limitations posed by the data distribution and model accuracy. Such limitation can be formulated as a bound on adversarial error rate, which captures the chance of success for an adversary to induce an incorrect outcome for QML models. 
This provides detailed insights into the minimum inherent adversarial vulnerability in QML and informs us on more effective defense mechanisms to reach these limits. The proliferation of QML technologies requires a security guarantee because, in the future, QML may be used in sensitive applications.

\textbf{Contributions.}
In this work, we present a novel, computable lower bound on the adversarial error rate, which provides a practical reference for model performance in adversarial attacks regardless of the quantum model's architecture. Our algorithm addresses the classical perturbation attack prevalent in the literature and the quantum perturbation attacks unique to the quantum model. We devise and test an example of such an attack based on the Projected Gradient Descent (PGD) \cite{Madry2017TowardsAttacks}. In Sec.\;\ref{sec:result}, we validate the effectiveness of the bound by comparing the derived bounds with the actual adversarial error rates observed in quantum models. The experimental results demonstrate our proposed bounds' practical effectiveness and applicability in real-world scenarios.
Overall, the key contributions are the following:
\begin{itemize}
  \item \textbf{Lower bound estimation method for QML}: 
  We present a new algorithm for quantum scenarios inspired by classical adversarial risk bound estimation methods~\cite{MahloujifarEmpiricallyRobustness}. Our approach is model-independent and is capable of addressing the unique quantum challenge of quantum perturbation attacks. Utilizing parallel computing, the algorithm efficiently computes the bound on an adversarial error and informs us of the limit of adversarial performance for any possible quantum models. 
  
  \item \textbf{Evaluation of practical effectiveness}: Our bound estimation algorithms are investigated against empirical benchmark data, showing a strong correlation between derived bound and observed adversarial error rates in quantum models. The results, detailed in Sec.\;\ref{sec:result}, confirm the practical applicability of our proposed bounds through benchmark scenarios.
  
  \item \textbf{Novel quantum attack strategy}: We have developed a new quantum attack strategy in the attack scenario of quantum perturbation attack. It is based on the Projected Gradient Descent (PGD) \cite{Madry2017TowardsAttacks}, a strong and widely-studied gradient-based attack in CML.

\end{itemize}

\section{Background and Literature review}
\subsection{Challenges in Quantum Machine Learning}
Quantum machine learning (QML) is an emerging field first proposed in 2019 \cite{Havlicek2019SupervisedSpaces, Sim2019ExpressibilityAlgorithms}. It integrates concepts from classical machine learning (CML) with developing quantum computing technologies. Unlike CML, which relies heavily on classical computing and is driven by extensive trial-and-error experimentation \cite{Domingos2012ALearning, Halevy2009TheData, Lecun2015DeepLearning}, QML is expected to rely mostly on the development of quantum computing technologies. However, the current stage of quantum technology, primarily Noisy-Intermediate-Scale Quantum (NISQ) devices, presents challenges for establishing far-reaching QML results across different quantum hardware with a wide range of capabilities and limitations \cite{Divya2021QuantumAlgorithms, Zhao2021AdvancedDevices, Bowles2024BetterModels, Innocenti2023PotentialMachines}.

Apart from the challenges posed by the vastly different quantum hardware in research, there is also a significant gap between the simulation of quantum models and the physical implementation of them. This further exacerbates the challenge of producing far-reaching results via experimentation similar to the CML approach.
Various simulated QML models have been tested, including supervised learning \cite{West2023ReflectionClassification, Senokosov2024QuantumClassification, Schatzki2021EntangledLearning}, unsupervised learning \cite{Okada2023ClassicallyPhases, KumarTiwari2024PotentialProblems, Park2022VariationalTransfer}, and reinforcement learning \cite{Dunjko2018AdvancesLearning}. Though often utilizing impractical oracle gates such as $n$-qubit amplitude encoding and $n$-qubit universal gate and assuming negligible noise, these models show promise in solving complex problems as quantum computing quality and scale improve. On the other hand, some experiments are also conducted on quantum hardware \cite{StroblReconstructingPhysics, Cao2024UnveilingAlgorithms}. Compared to the simulated works, these experiments are generally less complex due to limited qubits and high noise levels. These preliminary researches highlight a gap between the future practical quantum model and what we currently can analyze or experiment with. Thus, in our work, we adopt a less experimental approach and investigate the data structure and the allowed robustness for any potential model learning from the data. 
Even when facing these challenges, our results enable guidance in QML.


\subsection{Adversarial Quantum Machine Learning}
Many security challenges that CML faces stem from the models' need for trust in the data sources, lack of interpretability, and intrinsic vulnerability to unexpected and deliberate adversarial manipulation. As QML seeks to leverage quantum advantages within the framework of machine learning, it still inherits these classical challenges. Therefore, the security issues faced in CML could transfer or even exacerbate in QML environments. Fortunately, many theories and methods can be transferred or adapted from (classical) adversarial machine learning to QML. For example, the phenomena where a moderate increase in data dimension may be detrimental to the adversarial performance has been analyzed in the QML settings \cite{Liu2020VulnerabilityPerturbations, Gong2021UniversalClassifiers}, and the classical method of adversarial training has been introduced to QML \cite{Lu2020QuantumLearning, Liu2020VulnerabilityPerturbations, Banchi2022RobustLearning, Gong2021UniversalClassifiers}.

In the field of adversarial QML, one fundamental and popular problem category is the evasion attack and its defense, where adversaries attempt to stealthily induce incorrect decisions by exploiting the model's weakness \cite{West2023BenchmarkingScale, West2024DrasticLearning, Franco2024PredominantReview}. For classification problems, we expect a similar adaptation and knowledge transfer. For classification tasks in CML, advancements in machine learning models have enabled high-accuracy predictions of class labels. However, the accuracy can drop significantly when test samples are perturbed by adversaries \cite{Madry2017TowardsAttacks}. While heuristic defenses against adversarial attacks have been developed, many remain vulnerable to newer, adaptive attacks \cite{Athalye2018ObfuscatedExamples, Carlini2016TowardsNetworks}. For quantum machine learning development with large and deeper layers of quantum circuits and a higher number of qubits, we also expect a steadily increasing accuracy in non-adversarial settings and a long-lasting competition between attack and defense strategies \cite{Dosovitskiy2020AnScale}.

Quantum machine learning also brings many unique components to consideration other than using a quantum circuit mainly to process information. One important distinct component is the ``interface between quantum and classical information''. Examples of such interfaces include the encoding stage, where classical information is transferred to quantum states, and the measurement stage, where the classical information is extracted from the quantum states. These alterations of the classical machine learning formula introduce more opportunities to find and exploit vulnerabilities \cite{Gil-Fuster2023UnderstandingGeneralization}.

As quantum computing hardware progresses through the stage of noisy intermediate-scale quantum devices, the scope of research for QML is currently limited. However, there is significant potential for growth or QML \cite{Melnikov2023QuantumEngineering, Huang2022QuantumExperiments}, and research in adversarial QML is laying the necessary groundwork for future practical applications. In this emerging and evolving field of quantum adversarial machine learning, our paper pursues the theoretical and computable limits of model robustness in opposition to individual attacks or defenses. Being model-independent, the limit is a reference value for both current and future quantum models to compare regarding adversarial performance. It also provides insight into the inherent vulnerability of classical or (encoded) quantum data, depending on the use case. Thus, the bound is a benchmarking reference that contextualizes the model's robustness within the theoretical limit allowed by the given data distribution and enables evaluating model robustness by providing a theoretical limitation for reducing adversarial disruption via various defense strategies.

\subsection{Adversarial Attacks to QML}\label{sec:Atk Scen}
\begin{figure}
    \centering
    \includegraphics[width=0.9\linewidth]{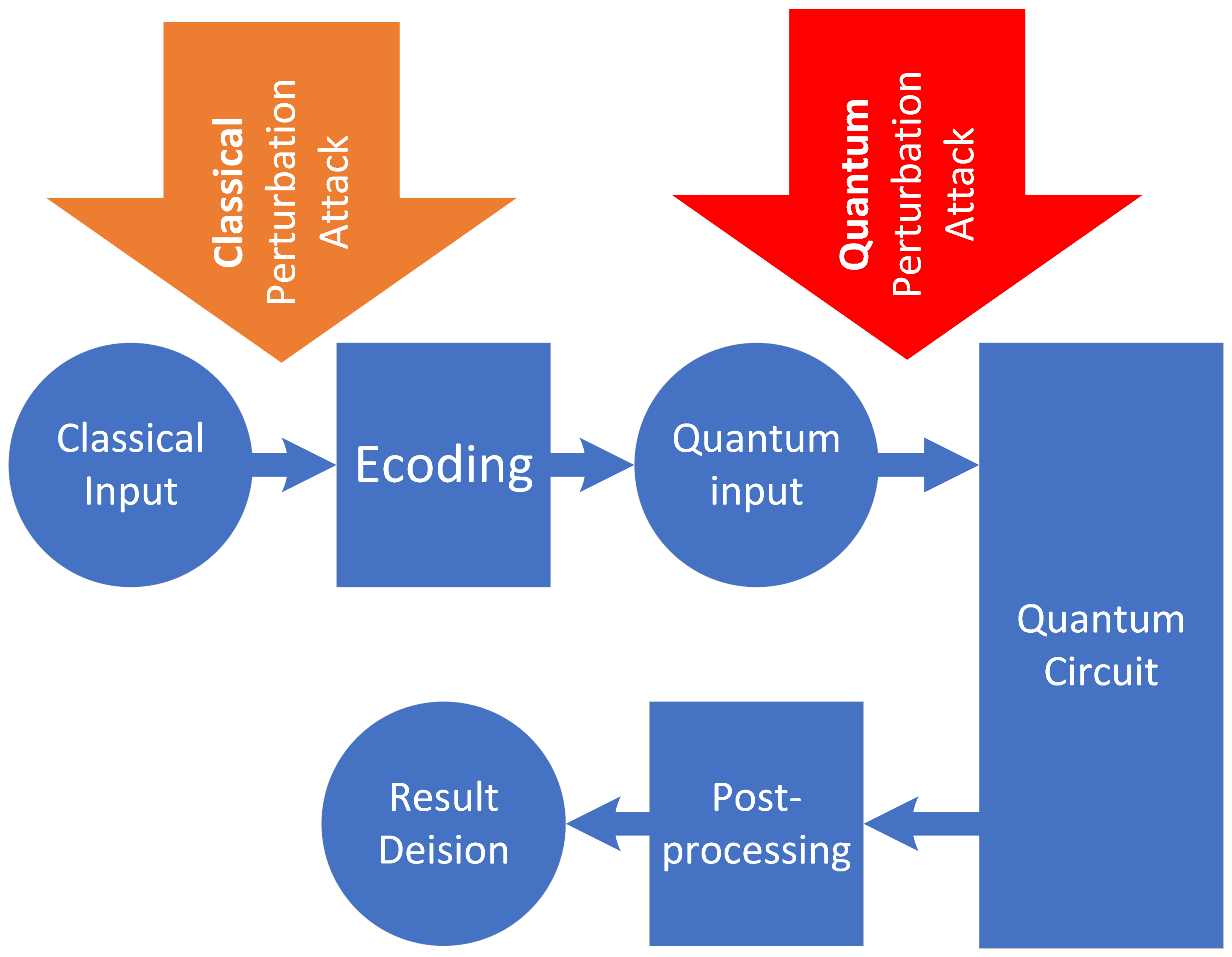}
    \caption{Illustration of classical and quantum input perturbation attack for QML when considering classical data as input.}
    \label{fig:Qattack}
\end{figure}

In the field of (classical) adversarial machine learning, a broad spectrum of attacks and defenses are explored. However, in Quantum Machine Learning (QML), research predominantly revolves around evasion attacks and corresponding defensive strategies \cite{Lu2020QuantumLearning, Ren2022ExperimentalQubits, West2023BenchmarkingScale}. In our work, we align our attack scenarios with the established frameworks of prior research. This alignment ensures that our methodology not only contributes to a deeper understanding of existing results but also aids in guiding future research that seeks to build upon these frameworks. We shall give a rigorous definition of these evasion attacks in Sec.\;\ref{sec:bound}. The difference between the quantum encoding stage quantum perturbation attack and the classical input perturbation attack is shown in Fig.\;\ref{fig:Qattack}

In the context of an evasion attack within a quantum setup, there is a unique step: the encoding of information onto the quantum computer. This step is bypassed when the input is inherently quantum. The attacker might target the system before the encoding of classical information, resulting in a conventional attack scenario restricted to a perturbation strength $\epsilon$ under classical distance metrics, such as $l_{\infty}$, $l_2$ and $l_1$ distances \cite{Lu2020QuantumLearning, Ren2022ExperimentalQubits, West2023BenchmarkingScale}. Conversely, when the attacker gains access to the encoded quantum state or the model's inputs are quantum states, an attack that alters quantum states under a specific quantum distance metric must be considered. The reason is that to detect the perturbation in the quantum states, quantum methods such as state verification need to be considered, where another distance metric shall define the differences between perturbed and unperturbed examples.

Currently, altering quantum states as a new attack vector seems relatively impractical, given that quantum computers are predominantly closed systems managed by large organizations. However, as quantum computing technologies evolve towards open-access and interconnected networks of quantum computers \cite{Simon2017TowardsNetwork, Fang2023QuantumPractice}, this scenario is set to change. In the era of quantum networks, quantum data will be distributed across these networks, potentially requiring state preparation or quantum data retrieval to occur outside the local, trusted quantum computing environments.

While existing literature has addressed quantum perturbations from environmental factors such as thermal noises \cite{Huang2021Information-TheoreticLearning,Lohani2022Data-CentricScience}, there is a noticeable gap in studies concerning deliberate or adversarial perturbations. The literature in adversarial quantum adversarial samples has not matched the depth and diversity of research on classical adversarial examples and perturbations. In our research, we aim to address both classical and quantum perturbations to ensure that our methodology and results remain relevant for near-term applications and future scenarios that leverage the distribution of quantum state and data.

\subsection{Bounds on Adversarial Errors}
\begin{figure}
    \centering
    \includegraphics[width=\linewidth]{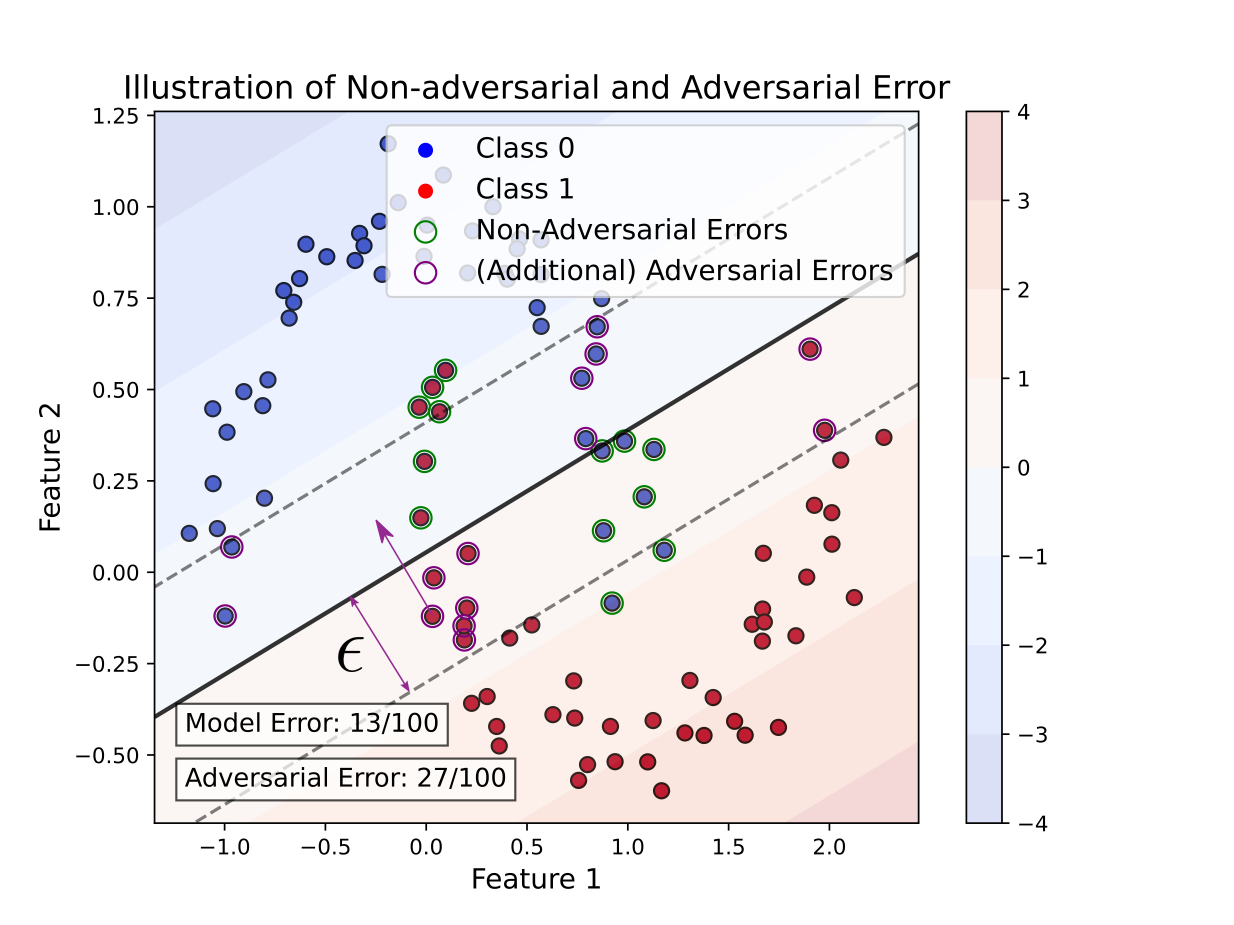}
    \caption{\textbf{An illustration of non-adversarial error and adversarial error}. The machine learning problem here is classifying the two-moon dataset with SVM. The solid line represents the decision boundary (chosen to be linear for simplicity). Samples within the green circles cause non-adversarial errors. Samples within the purple circles contribute to adversarial errors. With a gradient-based attack of strength $\epsilon$, these additional adversarial samples will be perturbed across the boundary, as shown by the purple arrow.}
    \label{fig:error_illu}
\end{figure}

In QML literature, the adversarial error (rate), the error rate of a model under an adversarial attack, is the most commonly used performance metric. The rigorous definition of non-adversarial error rate and adversarial error rate will be discussed in Sect.\;\ref{sec:errors}. To help understand the concepts, we illustrate the non-adversarial and adversarial error in a toy machine learning problem shown in Fig.\;\ref{fig:error_illu}. Providing a bound on such a performance metric can be very beneficial for developing and evaluating more robust models, not only QML but also CML. This is evidenced by the fact that, within the context of CML, existing research has explored the limit of adversarial performance under a variety of assumptions on the data distribution \cite{Gilmer2018AdversarialSpheres, Bhagoji2019LowerTransport, Mahloujifar2018TheMeasure, Fawzi2018AdversarialClassifier, Shafahi2018AreInevitable}. One of the concepts abstracted from adversarial error serving the same purpose of quantifying the possible performance of the models is \textit{adversarial risk}. Notably, \cite{MahloujifarEmpiricallyRobustness} proposed a practical method to evaluate this lower bound by sampling from datasets and comparing the estimated bounds with the best-known robust models at the time. This approach bridges the gap between theoretical robustness and experimental adversarial performance, providing a tangible reference for machine learning engineers in classical machine learning. In Sec.\;\ref{sec:bound}, we shall explain the connection between the practical adversarial error and the bound on adversarial risk in detail.



The bound also has direct implications for current experimental results. Consider an experimental implementation or simulation of a QML model that produces a specific adversarial error rate against a near-optimal attack. We can ascertain the quantum classifier's adversarial performance by comparing this test error rate with the estimated lower bound of adversarial risk. A test error rate close to the lower bound indicates that the quantum classifier minimizes the impact of adversarial perturbations to the theoretical limit allowed by the data distribution. In subsequent Sec.\;\ref{sec:bound}, we discuss the formalism of the adversarial risk lower bound and show why it is possible to compare the lower bound on adversarial risk with adversarial error.

\section{A Theoretical Bound on QML Adversarial Error for Evasion Attacks}\label{sec:bound}
In this section, we investigate the limits of adversarial QML model performance by defining a theoretical lower bound on adversarial risk, which converges to adversarial error when certain conditions are met.
By developing an empirical method to estimate the bound, we bridge the gap between machine learning theorists and practitioners. We initially evaluate the adversarial error rate, i.e., the effectiveness of the attack, which depends on the model and the attack. By the section's end, we derive a computable lower bound on adversarial risk independent of specific models and attacks.

\subsection{Adversarial Attack and Error}
\label{sec:errors}
In this section, we will provide a universal formalism of adversarial error that applies to both the classical perturbation and quantum perturbation attacks. We shall first consider classical perturbation. Suppose we have input samples and their corresponding labels, $(x_i,y_i)$ drawn from the testing set $\Omega_\textrm{test}$, i.e., $(x_i,y_i) \in \Omega_\textrm{test}$. For the classical scenario, we have d-dimensional input, $x_i \in \mathcal{X}$, where $\mathcal{X} = \mathbb{R}^d$, and the given labels $y_i \in \mathcal{Y}$, where $\mathcal{Y} = \{1,..,C\}$ with $C$ being the number of classes. Now we can construct the classifier $f$ as follows:
\begin{equation}
     f\; : \; \mathcal{X} \rightarrow \mathcal{Y}\;,
\end{equation}
a mapping between the input and output domain.

For the quantum perturbation attack scenario, we replace the input domain $\mathcal{X} = \mathbb{R}^d$ with $\mathcal{X} = \mathcal{H}$, where $\mathcal{H}$ indicate a Hilbert space where the quantum states live. Thus, the classifier $f$ must classify the quantum state and return a classical label $y$. In the following evaluations, both scenarios will be considered simultaneously unless otherwise specified.

\begin{definition}[Adversarial Error Rate]
    \label{def:AdvErr}
     A classifier $f(x)$ is evaluated on a testing set $\Omega_\textrm{test} \equiv \{(x_i, y_i)\}$ where $x_i$ are the input and $y_i$ are the label. The perturbation attack $A$ (as a generator of perturbation) on the classifier $f$ constrained by attack strength $\epsilon\in \mathbb{R}$ generates the perturbed sample $x'$, i.e., $x' = A(f, x, \epsilon)$ and $||x'-x||<\epsilon$. Then, the adversarial error will be:
    \begin{equation}
    \label{eq:AdvErr}
        \text{\normalfont AdvErr}_{\epsilon, A,f} = \frac{ \#\{(x_i,y_i)\in \Omega_\textrm{test} \; | \; f(x_i')\neq y_i\}} {\#\Omega_\textrm{test}}\;,
    \end{equation}
    where $\#$ denotes the cardinality of a set. \footnote{We use here $\#$ to distinguish cardinality from distance metric \mbox{$||\cdot||$} later introduced in this section.}
\end{definition}

The adversarial error rate defined above is one of the most commonly used evaluation metrics for assessing adversarial performance under test-time evasion attacks. Another similar metric for adversarial performance is adversarial risk. However, there is a subtle difference between adversarial error rate and adversarial risk, for which we will provide a bound. We will explain this difference and the conditions under which they are approximately equivalent after formally introducing adversarial risk later in this section.

An optimal attack $A_\text{opt}$, which induces the largest adversarial error on the model, constrained by $\epsilon$ in some distance metric \mbox{$||\cdot||$} would need to explore all possible points within an $\epsilon$ neighborhood of any sample $x$. In the following analysis, we use \mbox{$||\cdot||$} to denote one chosen distance metric relevant to the application of the specific model, such as $l_n$ distances for classical data or trace distances for quantum states. We will further discuss the motivation and calculation of the trace distance in Sec.\;\ref{sec:Qadapt}. With one chosen distance metric \mbox{$||\cdot||$}, under the optimal attack, any vulnerable test sample $x$ satisfies:
\begin{equation}
    \exists \; x' \;\text{\normalfont where}\; ||x'-x||<\epsilon \;\text{\normalfont and}\; 
    f(x')\neq y,
\end{equation}
where $y$ is the ground truth for input sample $x$.

To evaluate the adversarial error theoretically, we need to consider a probability metric space $(\mathcal{X}, \mu, ||\cdot||)$, where $\mathcal{X}$ is the continuous sample space, $\mu$ is the probability measure and \mbox{$||\cdot||$} is the distance metric that we have explained above. The probability measure $\mu$ takes any set $S \subseteq \mathcal{X}$ , and outputs the probability measure of the set $S$, $\mu(S) \in [0,1]$. Assume all testing samples $x$ are drawn from the probability metric space. 

We define the random variable $A\in\{0, 1\}$ describing the binary event of whether the optimal attack induces an error for a sample $x$ drawn from the probability metric space. Similarly, the adversarial error rate of a testing set of $n$ samples independently drawn from the distribution can also be formulated as a random variable, which shall have the same expectation as the per-sample binary event $A$. In practice, for a large sample size $n$, we expect the adversarial error rate to converge to its expectation,
\begin{multline}
    E[\text{\normalfont AdvErr}_{\epsilon, A_{\textrm{opt}}, f}] = \mu(S) \;\; \textrm{given}
    \\
    S = \Bigl\{ x \in \mathcal{X}\;| \;\exists x' \;\text{\normalfont where}\; ||x'-x||<\epsilon \;\text{\normalfont and}
    \\
    f(x')\neq f^*(x)\Bigr\} \;,
    \label{eq:AdvErr}
\end{multline}
where $f^*(x)$ is the ground truth function providing test labels in the data domain $\mathcal{X}$. Following the ground truth, we could naturally generate a testing set used in Def.\;\ref{def:AdvErr} with $\forall (x,y) \in \Omega_\textrm{test},\; y = f^*(x) \; $. By definition, the optimality of the attack also implies that any other attack $A$ will induce less error to the classifier than the optimal attack.

\subsection{Adversarial Risk}
Now, to further our theoretical evaluation, we consider another similar performance metric other than the adversarial error for models under test-time evasion attack, the adversarial risk \cite{Bhagoji2019LowerTransport, MahloujifarEmpiricallyRobustness}, for which we will provide a bound:
\begin{definition}[Adversarial Risk]
Given the ground truth, $f^*$, and the probability metric space $(\mathcal{X}, \mu, ||\cdot||)$, the adversarial risk of classifier $f$ under perturbation of strength $\epsilon$ is
\begin{multline}
\label{eq:AdvRisk}
    \text{\normalfont AdvRisk}_{\epsilon, f} = \mu(S)\;\; \textrm{given} 
    \\
    S = \Bigl\{ x \in \mathcal{X}\;| \;\exists x' \;\text{\normalfont where}\; ||x'-x||<\epsilon \;\text{\normalfont and}
    \\
    f(x')\neq f^*(x')\Bigr\} \;.
\end{multline}
\label{def:AdvRisk}
\end{definition}
Comparing (\ref{eq:AdvErr}) for adversarial error and (\ref{eq:AdvRisk}) for adversarial risk, we notice that the conditions that identify a sample $x$ as vulnerable to adversarial perturbation of strength $\epsilon$ are slightly different. For the expectation of adversarial error in (\ref{eq:AdvErr}), we have $f(x') \neq f^*(x)$. For the adversarial risk in (\ref{eq:AdvRisk}), we have $f(x') \neq f^*(x')$. Theoretically, we see a gap between the expectation of adversarial error and adversarial risk. 

However, if a random sample $x$ from the probability metric space $(\mathcal{X}, \mu, ||\cdot||)$ has a probability of 1 to satisfy the following:
\begin{equation}
    \textrm{for} \;\forall\;x'\textrm{ where }||x'-x||<\epsilon,\ f^*(x) = f^*(x')\;,
    \label{eq:condition}
\end{equation}
we will have $E[\text{\normalfont AdvErr}_{\epsilon, A_{\textrm{opt}}, f}] = \text{\normalfont AdvRisk}_{\epsilon, f}$. Qualitatively, the higher the probability of condition (\ref{eq:condition}) being satisfied, the closer the two aforementioned objects are. While not guaranteed, this convergence provides valuable insights into the applicability and limitations of the computable bound we shall provide in Sec.\;\ref{sec:alg}. We also compare the bound with the experimental adversarial error rate in Sec.\;\ref{sec:result}.

Note that condition (\ref{eq:condition}) has been mentioned in \cite{MahloujifarEmpiricallyRobustness} as an assumption to resolve the contradiction between the accuracy-robustness trade-off \cite{Tsipras2018RobustnessAccuracy} and the adversarial risk bound. Here, we argue that the condition is necessary to compare the derived adversarial risk with experimental error. When the condition is satisfied, we may replace adversarial error with adversarial risk without approximation:
\begin{equation}
    E[\text{\normalfont AdvErr}_{\epsilon, f}] \equiv \textrm{AdvRisk}_{\epsilon,f}\;.
\end{equation}
Here, we have omitted the adversarial error's dependency
on the attack with the assumption of implementing the optimal attack, i.e., the worst-case attack for the defender.

The defined adversarial risk is the maximal average error rate under an optimal adversarial attack. Here, we also define the non-adversarial risk (error rate) $\alpha$ by setting $\epsilon$ to 0 following the notation in previous literature \cite{Bhagoji2019LowerTransport, MahloujifarEmpiricallyRobustness}:
\begin{definition}[Non-adversarial Risk (Error Rate)]
\label{def:error}
Given the ground truth, $f^*$, the non-adversarial error of a classifier $f(x)$ is
\begin{equation}
    \alpha = \mu(S) \;\; \textrm{given} \;\;
    S = \Bigl\{ x \in \mathcal{X}\;| \; f(x)\neq f^*(x) \Bigr\} \;.
\end{equation}
\end{definition}

For supervised learning, if we treat the given labels as a result of the underlying ground truth, the definition above is equivalent to the clean error rate given a very large sample number, i.e., the sum of all off-diagonal elements of the confusion matrix. Thus, in the following section, we will use non-adversarial risk and non-adversarial error rate interchangeably.

\subsection{Characterising Adversarial Error Region}
Minimizing the adversarial risk from (\ref{eq:AdvRisk}) is challenging. Fortunately, studies have demonstrated a different approach to evaluate (\ref{eq:AdvRisk}) by a change of variable from the classifier $f$ to the error region $\mathcal{E}$ defined by any $f$ \cite{Talagrand1995ConcentrationSpaces}:
\begin{equation}
    f\;\rightarrow\;\mathcal{E} = \Bigl\{ x \in \mathcal{X}\;| \; f(x)\neq f^*(x) \Bigr\} \;.
\end{equation}
We also define the expansion of the error region $\mathcal{E}$ as follows:
\begin{equation}
    \label{eq:expansion}
    \mathcal{E}_\epsilon = \Bigl\{ x \in \mathcal{X} \;|\; \exists x'\in \mathcal{E} \;\textrm{where}\; ||x'-x||<\epsilon \Bigr\}\;.
\end{equation}
After the change of variable, the adversarial risk is simply:
\begin{equation}
    \textrm{AdvRisk}_{\epsilon, \mathcal{E}} = \mu(\mathcal{E}_\epsilon)
    \label{eq:ErrorRegion}\;,
\end{equation}
where $\mathcal{E}_\epsilon$ denote an expansion of $\mathcal{E}$ by $\epsilon$. The intuition behind the expansion of the error region is that for any point $x$ within the expanded region, there will be a neighboring point $x'$ lying within the original region, which guarantees the existence of $x'$ being an adversarial sample perturbed from $x$.

By transforming (\ref{eq:AdvRisk}) to (\ref{eq:ErrorRegion}), we may evaluate the adversarial risk with only the shape and volume of the error region expansion $\mathcal{E}_\epsilon$. However, by considering this formalism, we cannot directly recover a practical classifier $f$ from the optimization procedure when we try to find the lower bound of the adversarial risk, as we will discuss later. Nonetheless, the bound remains useful as a reference for developing robust models.

To summarize, we have transformed the experimental adversarial error into adversarial risk by considering the optimal attack and assuming the robust ground truth. For the next step, we will minimize the adversarial risk over all possible error regions $\mathcal{E}$ and, thus, obtain the minimum adversarial risk for all classifiers. Given $\alpha$, the adversarial risk bound $c_\textrm{adv}$ will be the global minima of the following constrained non-convex optimization problem.
\begin{definition}[Minimizing Adv. Risk via Error Region]
Given the data distribution on input $x$ and a clean learning error rate $\alpha$. The problem of minimizing the adversarial risk by minimizing the error region expansion is as follows:
\begin{align}
    &\min_\mathcal{E} \mu(\mathcal{E}_\epsilon) \nonumber \\
    &\mu(\mathcal{E}) = \alpha\;.
    \label{eq:bound}
\end{align}
We denote the minima of the problem as $c_\textrm{adv}$.
\end{definition}
By solving the minimization problem above regarding an input data distribution, we obtain the minimum adversarial risk $c_\textrm{adv}$ given the input clean learning error rate $\alpha$ and attack strength $\epsilon$ for any of the possible classifiers $f$. This is because by minimizing over all possible error regions $\mathcal{E}(f) \in \textrm{Pow}(\mathcal{X})$, which are regions defined by $f$, we are also minimizing over all classifier $f$. Thus, the resulting adversarial risk bound, ${c_\textrm{adv}}$, is a universal reference for a new model to look up to regarding adversarial performance. A robust model should have adversarial error close to the bound $c_\textrm{adv}$ for an optimal attack in the white box evasion attack setting.

The optimal attack can be considered the most effective known attack. The near-optimality of Projected Gradient Descent Attack and AutoAttack for evaluating evasion attacks is evidenced by their theoretical foundations in optimization, consistently superior performance in empirical studies, widespread adoption in the research community, and ability to generate strong adversarial examples across various models and datasets, making them highly reliable benchmarks for assessing adversarial robustness \cite{Carlini2016TowardsNetworks, Croce2020RobustBench, Zhu2019EfficientScenario, Liu2023ReliableEnsembles}. Thus, we may compare the bound representing the best performance allowed by the data distribution with the adversarial error rate of a classifier under these near-optimal attacks. To acquire the bound, we need to solve (\ref{eq:bound}), a difficult non-convex optimization problem. In the following sections, we shall introduce the heuristic method to solve the above minimization problem in the context of QML.

So far, we have considered minimizing the adversarial risk when the attack strength $\epsilon$ is fixed. For an optimal attacker, the adversarial error increases monotonically with $\epsilon$. Thus, the solution to (\ref{eq:bound}),  $c_\textrm{adv}$, should also satisfy:
\begin{equation}
    \textrm{Given} \; \epsilon' > \epsilon \;,\;  \textrm{Pr}[x\in \Omega_\textrm{test}\;|\;x\in\mathcal{E}_{\epsilon'}] \geq c_\textrm{adv}\;.
    \label{eq:AdvBound}
\end{equation}
An alternative approach is considering the maximum $\epsilon$ when the adversarial risk is bounded from above \cite{Liao2021RobustLearning}. This provides the maximum tolerable attack strength $c_\epsilon$ given a certain error budget $\textrm{Pr}_0$:
\begin{equation}
    \label{eq:alt_bound}
    \textrm{Given} \; \textrm{Pr}[x\;|\;x\in\mathcal{E}_\epsilon] < \textrm{Pr}_0 \;,\;
    \epsilon < c_\epsilon\;.
\end{equation}
The two conditional statements are contrapositive of the other when we set $\epsilon$ to $c_\epsilon$ and $\textrm{Pr}_0$ to $c_\textrm{adv}$. Thus, it is sufficient to find either of the bounds $c_\textrm{adv}$ and $c_\epsilon$ when analyzing the robustness of models. In this work, we use the former definition for robustness via adversarial risk lower bound in (\ref{eq:AdvBound}).

\section{Efficient Algorithms for Computation of Bounds}\label{sec:alg}

In this section, we introduce the methods used to estimate the theoretical bounds outlined in (\ref{eq:bound}) for the first time to the quantum domain and enhance them. We divide our contributions into quantum-specific adaptations enabling the estimation of the bound for quantum perturbation attack and the general improvement that applies to classical and quantum attacks.

To facilitate the discussion, we explain the intuition behind the algorithm we will be diving into in this section. Equation (\ref{eq:bound}) highlights an optimization problem concerning both the adversarial and non-adversarial risk. We make an unbiased estimation of these risks by counting the number of samples in the estimated region $\hat{\mathcal{E}_\epsilon}$ and $\hat{\mathcal{E}}$. We then parameterize and limit the choice of $\hat{\mathcal{E}}$ to the union of $T$ hyper-spheres,
\begin{equation}
    \hat{\mathcal{E}} = \bigcup_i^T Sphere(c_i, r_i)\;.
\end{equation}
For the expansion of $\hat{\mathcal{E}}$ corresponds to attack strength $\epsilon$, we have
\begin{equation}
    \hat{\mathcal{E}_\epsilon} = \bigcup_i^T Sphere(c_i, r'_i(\epsilon))\;,
    \label{eq:ExpER}
\end{equation}
where $r'_i$ is the expanded radius of the sphere according to the attack strength $\epsilon$. For each step of the bound estimation algorithm, a sphere centered on a data sample is chosen to minimize the extent of adversarial disruption.

The algorithm employs hyper-spheres to incrementally carve out the error region from the entire data space of the training dataset \cite{MahloujifarEmpiricallyRobustness}. Each sphere fitting step, confined to available sample locations, poses a discrete optimization problem minimizing adversarial risk. The efficient execution of this subroutine is pivotal for the overall efficient computation. After obtaining an optimized region $\hat{\mathcal{E}}$ by fitting hyper-spheres, an evaluation of both the adversarial risk and non-adversarial risk on a testing set will produce an Adv. / Non-adv. risk pair $(c_a, c_{na})$ indicating that the experimental Adv. error rate of models with a clean error rate higher than $c_{na}$ should have Adv. error rate higher than $c_a$.

\subsection{Computing Adversarial Quantum ML Bound}\label{sec:Qadapt}
This section introduces the novel method that evaluates the adversarial QML bound for quantum perturbation attacks for the first time. The algorithm illustrated above requires the computation of pairwise distances and the expansion of hyper-spheres within the Hilbert Space where quantum states live. These adaptations are crucial for evaluating the bounds in the quantum attack scenarios discussed in Section.\;\ref{sec:Atk Scen}. 

\subsubsection{QUANTUM PERTURBATION ATTACK}
A quantum perturbation on the quantum input sample $\ket{\psi}$ generated by encoding the classical sample $x$ may or may not lay within the quantum data manifold of the encoded quantum states, the subspace of the Hilbert space where the encoding scheme span with arbitrary real input $x$. Therefore, there are no classical input samples corresponding to these perturbed quantum states and no ground truth in the classical sense for these perturbed inputs outside the quantum data manifold. In this work, we consider attacks within the data manifold. Thus, the perturbed quantum inputs are always pure states. 

\paragraph{Perturbation detection \label{sec:detec}}
Quantum perturbations introduced after the encoding stage directly modify the quantum state. Certification methods \cite{Zhu2019EfficientScenario, Pallister2018OptimalMeasurements} indicate that the number of necessary experimental runs to detect such perturbations scales inversely with the infidelity $\varepsilon$ between the intended input pure state $\ket{\psi}$ and the adversarial input mixed state $\sigma$:
\begin{equation}
\label{eq:infid}
    \varepsilon = 1 - \bra{\psi}\sigma\ket{\psi}\;.
\end{equation}
In this work, we only consider the adversarial inputs that are pure states consistent with the aforementioned in-manifold ground truth discussion. 

If $\sigma$ is a pure state, the infidelity can be simply expressed by the quantum trace distance \mbox{$||\cdot||_{QT}$}:
\begin{equation}
    \varepsilon(\ket{\psi}, \ket{\phi}) = \Bigl |\ket{\psi}\bra{\psi}-\ket{\phi}\bra{\phi} \Bigr |_{QT}\;.
\end{equation}
In the following discussion, we simplify the notation of infidelity and trace distance to $\varepsilon$ and $D_{QT}$. 

Motivated by the perturbation detection criteria, we define the quantum perturbation strength in this work by trace distance. 

Consider a fixed number of shots $n$ allowed for each input test sample to use for state certification, i.e., a finite verification resource. Given a fixed significance level, the states need to have infidelity of at least $\varepsilon = O(1/n)$ to be distinguishable from each other. We denote the corresponding trace distance threshold as $D_{QT} = \sqrt{\varepsilon} = O(n^{-\frac{1}{2}})$. For the attackers to evade such detection, they must have their perturbed state within the detection threshold $D_{QT}$. Thus, we have justified using the trace distance to measure attack strength when the defender detects perturbation with quantum state certification methods. This is similar to the classical $l_{\inf}$ distance in the sense that both of them imply the lack of distinguishability when the distance between the original and perturbed data point is small.

\paragraph{Expansion of quantum hyper-sphere for bound calculation}
The expansion of a region in the metric space defined by sample domain and distance metric ($\mathcal{X}$, \mbox{$||\cdot||$}) is given by (\ref{eq:expansion}). Here, we consider a special case of the error region where the error region is the union of spheres. In the classical ML case with the real vector space and $l_2$ distance, $(\mathbb{R}^n, ||\cdot||_{l_2})$, the expansion of the sphere centered on $c$, with radius $r$ is simply another sphere centered on $c$ with radius $r'=r+\epsilon$. For the quantum perturbation attack, we consider the metric space $(\mathcal{P}(\mathcal{H}), ||\cdot||_{\textrm{TD}})$ where $\mathcal{P}(\mathcal{H})$ is the set of general pure state in the Hilbert space $\mathcal{H}$ and $||\cdot||_{\textrm{TD}}$ is the trace distance.

For the quantum case, the radius of the expanded hyper-sphere will be $r'=r\sqrt{1-\epsilon^2} + \epsilon\sqrt{1-r^2}$ instead of the simple addition rule $r'=r+\epsilon$. The following theorem establishes the relationship between the radii of expanded hyper-spheres and itself in a Hilbert space with trace distance:
\begin{theorem}
     In the metric space $(\mathcal{P}(\mathcal{H}), ||\cdot||_{\textrm{TD}})$ of pure states and trace distance,
     \begin{equation}
         Sphere_{c,r}^\epsilon = Sphere_{c, r'}\;,
     \end{equation}
     where $Sphere_{c,r}^\epsilon$ is the expansion of the set defined by the surface and interior of a sphere, i.e., $Sphere_{c,r} = \{x\;|\;|x-c|\leq r\}$. $Sphere_{c, r'}$ is the sphere with the same centre $c$ but a larger radius $r'$, where 
     \begin{equation}
        r'=r\sqrt{1-\epsilon^2} + \epsilon\sqrt{1-r^2}\;.
        \label{eq:rprime}
     \end{equation}
     \label{the:1}
\end{theorem}

\begin{proof}
    Here we provide a constructive proof where we show $Sphere_{c,r}^\epsilon \subseteq Sphere_{c, r'}$ and $Sphere_{c,r}^\epsilon \supseteq Sphere_{c, r'}$ separately. In the proof below, we use the Dirac Bra-ket notation $\ket{v}$ to denote the state vector of the density matrix $v=\ket{v}\bra{v}$. The trace distance will become half of the matrix trace norm of the differences in the density matrix, i.e., $D_{1,2} = \frac{1}{2}||v_1-v_2||_1$.

    First, we introduce Bures angle \cite{Bures1969An-Algebras} $\theta_{v,u}\in[0,\pi/2]$ between two quantum states $\ket{u}$ and $\ket{v}$ \begin{equation}
    \cos \theta_{v,u} = |\braket{v}{u}|\;,
    \end{equation}
    Using the Brues angle, we have:
    \begin{equation}
        D_{1,2} = \sin \theta_{1,2}\;.
    \end{equation}

    Next, we substitute $r$, $\epsilon$ and $r'$ with the corresponding Brues angle $\theta \in [0,\pi/2]$: $r= \sin\theta_r$, $\epsilon= \sin\theta_\epsilon$,  $r'= \sin\theta_{r'}$. Now, we also see the intuition of the choice of $r'$: $\theta_{r'} = \theta_r + \theta_\epsilon$. Using the Bures angle, the problem becomes proving the two statements below,
    \begin{itemize}
        \item Given any quantum state $x$ and $y$ satisfying $ \theta_{c,y} \leq \theta_r$ and $\theta_{y,x} \leq \theta_\epsilon$, we have $\theta_{x,y}\leq \theta_r + \theta_\epsilon$.
        \item Given any quantum state $x$ satisfying $\theta_{x,c} \leq \theta_r + \theta_\epsilon$, there exist quantum state $y$ satisfying $\theta_{c,y}\leq \theta_r$ and $\theta_{y,x}\leq \theta_\epsilon$.
    \end{itemize}

    To prove the first statement, it is sufficient to prove the triangle inequality of angles in the pure state Hilbert space:
    \begin{equation}
        \theta_{v_1,v_2}+\theta_{v_2,v_3}\geq\theta_{v_1,v_3}\;,
    \end{equation}
    where $v_1, v_2, v_3$ are complex unit vectors. The equality is achieved if and only if the three vectors are linearly dependent and $\braket{v_1}{v_2} \braket{v_2}{v_3} \braket{v_3}{v_1}$ is a real number. To prove the inequality above, we utilize the non-negative determinant of the Gram matrix $G_{ij} = \braket{v_i}{v_j}$. We provide detailed proof in Appendix.\;\ref{app:tri}.
    
    We shall prove the second statement by giving an example for the quantum state $y$. First, we decompose the given state $\ket{x}$ in the subspace span by $\ket{c}$ and $\ket{x}$:
    \begin{equation}
        \ket{x} = \cos\theta_{x,c}\ket{c} + \sin \theta_{x,c}\ket{x_\perp}\;,
    \end{equation}
    where $\ket{x_\perp}$ and $\ket{c}$ ($\braket{x_\perp}{c} = 0$) form an orthonormal basis for the subspace. For $\theta_{x,c} \leq r$, the state $\ket{x}$ is inside the original sphere; thus, the proof is trivial. Below, we consider cases where $\theta_{x,c} > \theta_r$.
    
    To prove the existence of such quantum state $\ket{y}$, we give an instance of $\ket{y}$ satisfying $\theta_{c,y}\leq \theta_r$ and $\theta_{y,x}\leq \theta_\epsilon$:
    \begin{equation}
        \ket{y} = \cos\theta_r \ket{c} + \sin\theta_r \ket{x_\perp}\;.
    \end{equation}
    Using the $\ket{y}$ above, we have $\bra{c}\ket{y} = \cos\theta_r$ and thus, $\theta_{c,y} = \theta_r$. We will find $\theta_{y,x}$ by calculating $\bra{x}\ket{y}$.
    \begin{align}
        \bra{x}\ket{y} 
        &= \cos\theta_{x,c} \cos\theta_r + \sin \theta_{x,c} \sin\theta_r \notag \\
        \cos \theta_{x,y} 
        &= \cos (\theta_{x,c} - \theta_r)
    \end{align}
    As $\theta_{x,c} > \theta_r$ and $\theta_{x,c} \leq \theta_r + \theta_\epsilon$, we have
    \begin{equation}
        \cos \theta_{x,y} \geq \cos \theta_\epsilon\;.
    \end{equation}
    Thus, we arrive at $\theta_{x,y} \leq \theta_\epsilon$, completing the proof of the second statement.
    
    Now, we have proved the two statements corresponding to the two directions of set equality $Sphere_{c,r}^\epsilon = Sphere_{c, r'}$. Thus, we have proved Theorem \ref{the:1}. 
\end{proof}

With Theorem \ref{the:1}, we calculate the expanded radius for the bound evaluation of the quantum perturbation attack scenario using the formula (\ref{eq:rprime}): $r'=r\sqrt{1-\epsilon^2} + \epsilon\sqrt{1-r^2}$ in the theorem.

\subsubsection{COMPUTING PAIR-WISE TRACE DISTANCE \label{sec:pair-wise_distance}}
The bound estimation algorithm discussed above requires the computation of pair-wise trace distance between any two encoded quantum states. Fortunately, we do not need to compute the two encoded quantum states (potentially exponential in memory size on classical computers) for some basic encodings, such as Amplitude encoding and Phase encoding. In both cases, due to the mathematical simplification, the complexity of calculating a single pair is the same as for classical $l_n$ distances at $O(d)$, where $d$ is the data dimension. Notice that the fidelity (also infidelity) trace-distance relationship: $1-\varepsilon = 1-D_{QT}^2$, to compute the pair-wise trace distance, we will only provide two brief examples of calculating the fidelity metric in the following sections.

\textbf{Amplitude Encoding}
Amplitude encoding encodes the normalized classical data $\{ u_i = x_i/\sqrt{\sum_i x_i^2} \}$ vector to the amplitude of a quantum state. If the dimension of the data does not equal $2^n$, we pad the remaining complex amplitudes with zero. Thus, we have
\begin{equation}
    \ket{\psi} = \sum_{i=0}^{2^n-1} u_i \ket{i}\;.
\end{equation}
The fidelity between two encoded quantum states from classical input $x^{(1)}$ and $x^{(2)}$ will be
\begin{equation}
    F(\ket{\psi_1}, \ket{\psi_2}) = \Bigg[ \frac{ x^{(1)} \cdot x^{(2)}}{|x^{(1)}|_{l_2} |x^{(2)}|_{l_2}} \Bigg]^2
\end{equation}

\textbf{Angle Encoding}
Angle encoding uses single qubit gates applied on each qubit to encode one feature per qubit. The unitary operator of these encoding gates is
\begin{equation}
    U = \bigotimes_i^D R_x(\frac{\pi}{2}x_i)\;.
\end{equation}
Thus, we can calculate the fidelity between two encoded quantum states with classical input $x^{(1)}$ and $x^{(2)}$ as
\begin{align}
    F(\ket{\psi_1}, \ket{\psi_2}) &= \abs{\bra{0}U^\dagger_{x^{(1)}} U_{x^{(2)}} \ket{0}}^2\\
    &= \abs{\bra{0}\bigotimes_i^D R_x[\frac{\pi}{2}(x_i^{(1)}- x_i^{(2)})]\ket{0}}^2\\
    &= \prod_i^D \textrm{cos}^2[\frac{\pi}{2}(x_i^{(1)}- x_i^{(2)})]
\end{align}

\subsection{PARALLELIZED BOUND CALCULATION}
Apart from the adaptations required to calculate the bound for the quantum attack scenario, our algorithms achieve high efficiency and accuracy in the bound estimation procedure. In this section, we will first discuss how we improve the speed of computing the bound by using a parallelized subroutine that can be executed more efficiently on modern computing hardware such as CPU clusters or GPUs. Then, we will discuss how adding a regression procedure at the end of the bound evaluation process improves the accuracy of the bound.

From a computational perspective, this approach leverages the power of parallel computing (unlike \cite{MahloujifarEmpiricallyRobustness}). Our algorithm employs strategic precomputation of pair-wise distances and k-nearest neighbors, enabling the majority of processing to run on a GPU (Graphics Processing Unit). This innovative GPU-accelerated design drastically reduces computation time while maintaining high-quality results and ensures the method is scalable for large and complex datasets.

We also incorporate a linear regression procedure to analyze the relationship between adversarial and non-adversarial errors. This novel approach ensures that the algorithm's output accurately reflects the defined error region $\mathcal{E}$, where $\mu(\mathcal{E})=\alpha$, as specified in (\ref{eq:bound}).

\subsubsection{PARALLELIZED SUBROUTINE}
Our algorithm detailed in Appendix \ref{app:alg} is highly parallelized and can be run on conventional GPU. It achieved a comparable complexity scaling to the first algorithm of adversarial risk lower bound estimation \cite{MahloujifarEmpiricallyRobustness}. The algorithm we propose utilizes the matrices efficiently stored on conventional RAMs instead of the ball tree in the original algorithm to store the pair-wise distances $D_{ij} = ||x_i - x_j||$. The algorithm procedure is outlined below:
\\
\small
\begin{enumerate}[label=Step \arabic*), start=0, leftmargin=4em]
    \item Compute the Pair-wide distance $D_{ij}$.
    \item Sort the distances to find sorted distances $D^{(s)}$ and index matrix $I$.
    \item Choose a centre $c$ and radius $r$ to fit the sphere, the range of possible radius depends on T and the remaining budget of non-adversarial risk.
    \item Find the increase of the non-adversarial risk region approximated by the sample portion included in the region.
    \item Find the increase of the adversarial risk region utilizing the sorted distances $D^{(s)}$ and bisection method.
    \item Repeat Steps (2) and (4) for all combinations of centers and radii and find the minimal extra risk, i.e., $\{c', r'\} = \min_{c, r} \textrm{AdvRisk}_{c,r} - \textrm{Risk}_{c,r}$. This sphere will be chosen as the optimal sphere for this step.
    \item Trim the sorted distances $D^{(s)}$ based on $I$ to account for the points already included in adversarial and non-adversarial risk regions.
    \item repeat step (2) to (6) $T$ times to obtain the full adversarial and non-adversarial risk region.
\end{enumerate}
\normalsize

After sorting $D_{ij}$ along the $j$ axis in Step 1, we generate and store the sorted list $D^{(s)}_{ik}$ and the index matrix $I$. By processing $D^{(s)}_{ik}$ and $I$ during the sphere-fitting procedure, we keep track of both the nearest distance rank $k$ and the original index $(i,j)$. Thus, in Step 4, given the not-expanded hypersphere including $k$ samples representing the error region $\mathcal{E}$, we can efficiently compute the set of samples included in an expanded hypersphere representing the expanded error region $\mathcal{E}_\epsilon$ by searching the sorted ascending list of distances $D^{(s)}_{ik}$ for all potential spheres centered on the $i$th sample.
We emphasize that the iteration through both the centers $i$ and the error region size $k$ can be executed in parallel, making the algorithm scalable for large datasets with many samples. Additionally, searching the sorted list is efficiently handled by packages such as PyTorch and CuPy using either CPU or GPU. Overall, the algorithm's complexity is dominated by Step 4, which finds the adversarial risk for a given sphere. For a single iteration of Step 4, the complexity is $O(\log n)$. Thus, the overall complexity is $O(n^2 \log n)$, where $n$ is the number of samples. 
It is worth noting that the complexity of the preprocessing step is $O(n^2 d)$; however, due to the large constant associated with the main loops, the total runtime is dominated by the main loop at $O(n^2 \log n)$.

Appendix\;\ref{app:alg} provides a more detailed complexity analysis.


\subsubsection{BOUND ESTIMATION THROUGH REGRESSION}
The algorithm is designed to minimize adversarial risk within a given dataset (training set). To ensure an unbiased estimate of the theoretical bounds, it is essential to partition the dataset into training and testing sets, mirroring traditional machine learning practices. We output the adversarial risk estimated on the testing set as the estimated bound. This division typically results in a difference in both non-adversarial and adversarial risks between the training and testing sets. This difference means that the output test non-adversarial risk will not equal the input $\alpha$ of Algorithm.\;\ref{alg:loops}. To rectify this divergence, a regression procedure is employed to estimate the bound value accurately. Assuming that changes in $\alpha$ linearly affect the bound, we can interpolate the testing set adversarial risk value at exactly input $\alpha$ through a series of bounds results initialized with different training set non-adversarial risks. The details on and the implementation of the full bound estimation algorithms via regression are discussed in Appendix \ref{app:alg}. Here, we briefly explain the regression procedure and the hyper-parameters involved.

\small
\begin{enumerate}[label=Step \arabic*), start=0, leftmargin=4em]
    \item Initiates a list of target non-adversarial risk $\{\alpha_{\nu}\}$ close to the actual error rate of the model that we evaluate the bound against.
    \item Partition the entire data set randomly to obtain a training set and a testing set.
    \item Run the bound finding algorithms on the training set with target error rate $\alpha_{\nu}$ and estimate the adversarial risk $AdvRisk_{\nu}$ and non-adversarial risk $Risk_{\nu}$ on the testing set.
    \item Repeat Step 1) and Step 2) for all the $\alpha_{\nu}$.
    \item Run linear regression on the relationship $AdvRisk_{\nu}$ vs the non-adversarial risk (error rate) $Risk_{\nu}$.
    \item Use the regression result to interpolate the minimum adversarial risk at the actual error rate of the model that we evaluate the bound against.
\end{enumerate}
\normalsize

To initiate the list of target non-adversarial risk $\{\alpha_{\nu}\}$, we introduce three hyper-parameters, $m$, $\alpha_l$, and $\alpha_u$, to manage the iterations over different $\alpha_{\nu}$ values. $\alpha_l$, and $\alpha_u$ specify the range of non-adversarial risk while $m$ specify the total number of samples within this range. Increasing $m$ results in more experiments with random partitions of $S$. Increasing the range $[\alpha_l, \alpha_u]$ enhances the coverage of the $\alpha$ and decreases the chance of $\alpha$ falling outside the observed values of the testing set non-adversarial risk. However, a broader range will likely reduce the linearity assumed for the regression procedure due to a greater spread around the local $\alpha$. Thus, we adjust the $\alpha$ range and choose a large enough $m$ before running the algorithm to obtain the best result.


\section{EXPERIMENTS AND RESULTS}\label{sec:result}

In this section, we demonstrate the utility of the quantum adversarial lower bound by applying the algorithm to a realistic data set. We observe that the bound we have derived and estimated sits close to the experimental adversarial error rate for some instances of the model, whereas further for others. By simulating quantum computers using the pennylane \cite{Bergholm2018PennyLane:Computations} package, we train multiple models on the MNIST and FMNIST datasets and attack them with adversarial attacks for both classical and quantum perturbation attack scenarios. We also compute the estimated bound value for the given dataset, the non-adversarial error rate, and attack strength. Then, we compare the estimated bounds with the adversarial error rates. 

\subsection{EXPERIMENT SETUP}\label{sec:setups}

\paragraph{Model training}
In this demonstration of the bound, we consider image classification tasks on the MNIST and FMNIST datasets. The structure of the quantum models follows the formalism of the Quantum Variational Circuit (QVC)\cite{Cerezo2020VariationalAlgorithms, Anschuetz2022QuantumTraps}. The model consists of three simple stages: encoding, quantum circuits, and a simple classic post-processing stage.

\begin{figure*}
    \centering
    \includegraphics[width=0.6\linewidth]{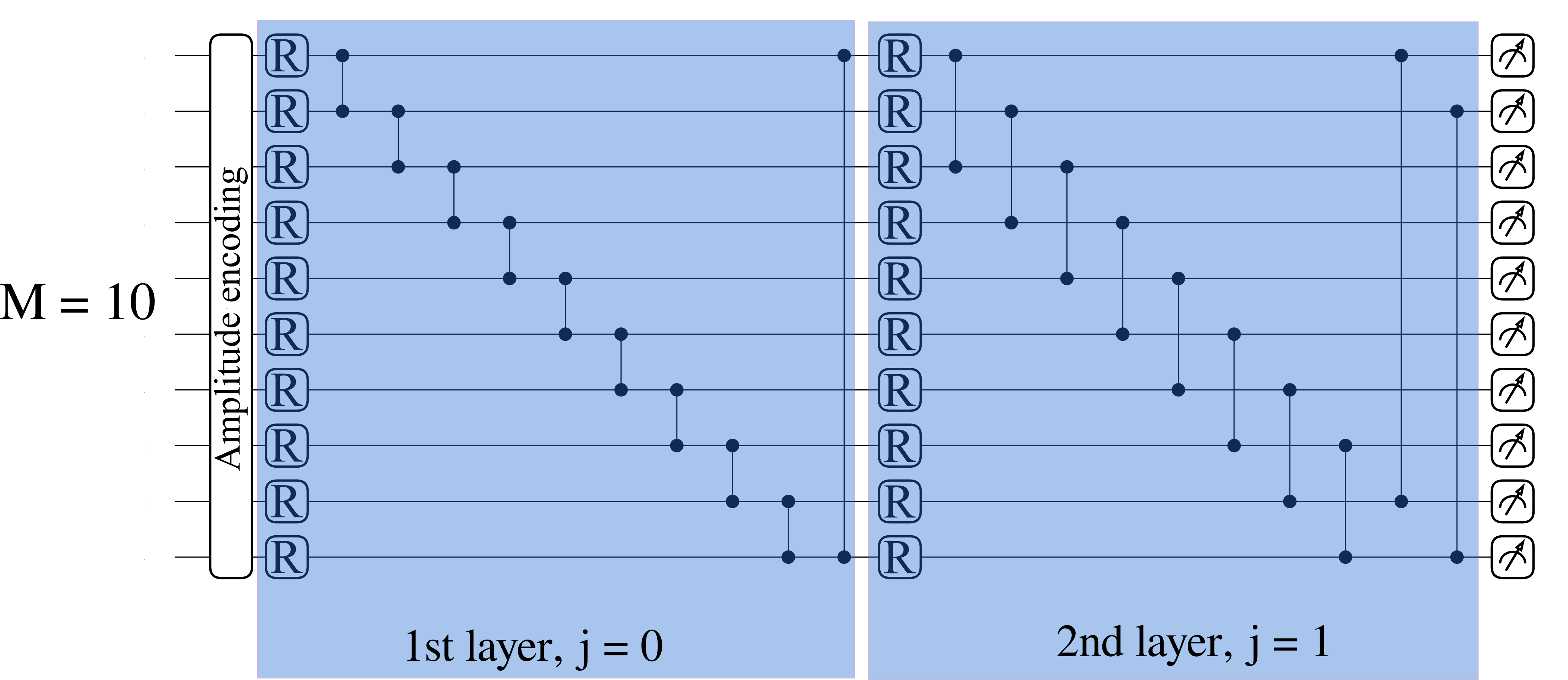}
    \caption{\textbf{Schematic diagram of encoding and quantum circuit step.} From the left to the right, the quantum circuit consists of an input layer, ansatz, and measurement gates. The input layer is the amplitude encoding procedure. The ansatz consists of a pattern of parameterized quantum gates. Two such layers are shown here. The single qubit gates denoted by ``R'' are general $u_3$ gates with 3 parameters. In the $j$th layer \{($i$, $i+j \mod M$)\} qubits are entangled by CZ gate. We use 200 layers of these gates for all the models in the result section. At the end of the circuit, we measure and output the expectation $\langle \sigma_z^{(i)} \rangle$ for each of the 10 qubits corresponding to the 10 classes.}
    \label{fig:qc}
\end{figure*}

The encoding and quantum circuit stages, as shown in Fig.\;\ref{fig:qc}, should be implemented on a 10-qubit quantum computer. However, in our machine learning experiment, we simulate the quantum components in this model to avoid the higher noise level in current quantum hardware and probe the model's performance with a circuit depth much larger than what is allowed on current hardware. For the encoding stage, we adopted amplitude encoding mentioned in Sect.:\ref{sec:pair-wise_distance}:
\begin{equation}
    \ket{\psi} = \sum_{i=0}^{2^n-1} u_i \ket{i}\;.
\end{equation}
For the quantum circuit stage, the ansatz of the quantum layer is Pennylane \cite{Bergholm2018PennyLane:Computations} \texttt{StrongEntanglingLayers} with CZ gate as the two-qubit gates. At the end, the circuit output 10 features by reading the measurement output and estimating the expectation value of observable $\sigma_z$ for each qubit at the end, i.e., $\langle\sigma_z^{(i)}\rangle$. Thus, the output of the quantum circuit stage is a 10-dimensional vector with each element $\langle\sigma_z^{(i)}\rangle \in [-1,1]$. Note that among the three stages, only the quantum circuit stage is trainable, with quantum rotational gate parameters as the trainable parameters.

For the final stage, the classical post-processing stage, it aims to introduce some non-linearity to the model and ensure the model predicts output with high confidence. We use a softmax layer for training, and during testing, we use the argmax layer to make deterministic decisions. Note that this formulation of quantum models is not the only possible formulation. Our estimated bound given the same input should apply to any other model for both classical and quantum attack strength\footnote{To calculate the quantum version of the bound with a quantum attack strength, pair-wise trace distances between encoded samples are required, thus, the models have to use the same encoding scheme for it to share the same estimated bound}.

We consider different softmax temperatures $t$ for the softmax function to generate a range of instances of the model with distinct error and adversarial error rates. The softmax temperature defined as follows:
\begin{equation}
    \texttt{Softmax}(Z_i) = \frac{e^{Z_i/t}}{\sum_j e^{Z_j/t}}
\label{eq:softmax}
\end{equation}
The temperature $t$ affects the `smoothness' of the \texttt{Softmax} function. When the temperature $t$ decreases, the \texttt{Softmax} function will become sharper and more similar to an \texttt{Argmax} function. Note that classical machine learning models do not usually tune the softmax temperature. The reason is that the linear layer before the \texttt{Softmax} function serves a similar role as the temperature $t$, which scales the input features to the \texttt{Softmax} function up or down by a constant multiplier.

Intuitively, to best approximate the testing model, we would like to have a \texttt{Softmax} function as close to the \texttt{Argmax} function as possible with low temperature. However, when decreasing the temperature, we observe that the expected quantum output $\textrm{E}[Z_i]$ of a new test sample becomes closer to one another. We argue that as the temperature decreases and the softmax function becomes sharper, the demand on the quantum circuit during training to produce sharper and more distinguishable output decreases, thus rendering the 10 outputs of the quantum circuit closer to each other. 

This lack of distinguishability for models trained with low-temperature softmax function requires greater precision on the estimations of $\textrm{E}[Z_i]$ during testing or model deployment so that the final decision on the class label is statistically stable. This precision requirement will increase the number of quantum circuit runs to generate estimations of $\textrm{E}[Z_i]$. In our work, we perform simulations of the quantum circuit that directly obtains the precise value of $\textrm{E}[Z_i]$, so the aforementioned issue does not apply. In practice, a higher number of quantum circuit runs can impact impact the evaluation and training speed of the quantum machine learning model. 

We have done preliminary testing adversarial training with the PGD attack and the attack we devised, which will be discussed in the following section. We find that the robustness benefits from conventional adversarial training are negligible, consistent with \cite{West2023BenchmarkingScale}. Thus, we choose not to incorporate adversarial training in this work. However, the detailed results of these preliminary tests shall not be included in this work.


\begin{figure}
    \centering
    \includegraphics[width=1\linewidth]{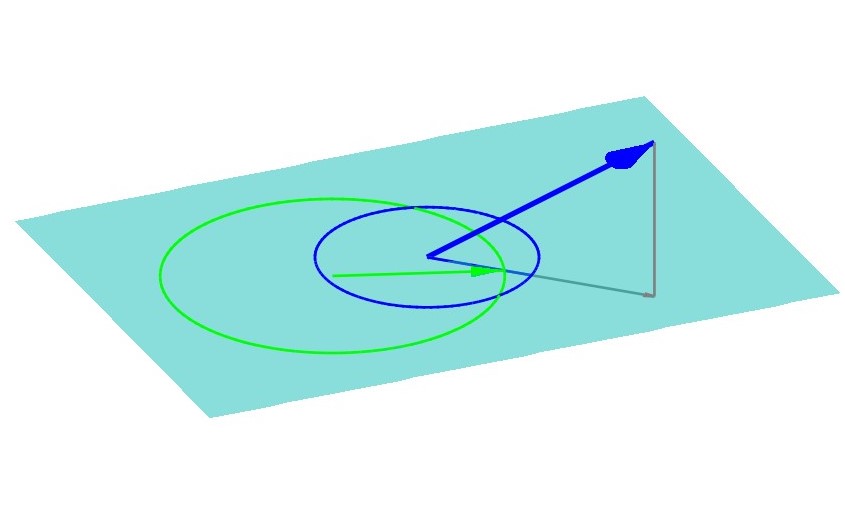}
    \caption{\textbf{An illustration of the TD-PGD attack for the amplitude encoding.} For visual clarity, instead of using a spherical surface, we denote the spherical normalized vector space of input real quantum amplitude with a flat surface. Step \ding{172} projects the gradient (blue arrow) to the normalized vector space and assigns an amplitude of step-wise attack strength (blue circle). Step \ding{173} clip and project the step-wise perturbation to the overall attack strength (green circle) if necessary and update the attack (green arrow).}
    \label{fig:TD_PGD}
\end{figure}

\paragraph{Classical and Quantum Perturbation Attack}
For the classical $l_2$ attack, we apply the Projected Gradient Descent attack to the model. The gradient of the model is obtained directly by simulated backpropagation. In practice, quantum-specific methods such as parameter-shift \cite{Mitarai2018QuantumLearning, Wiersema2023HereGradients} will be required to propagate the gradient through the quantum circuit. For the quantum perturbation attack under the attack scenario we introduce, the quantum adversarial sample state needs to be generated by the same encoding scheme and evade detections based on quantum state verification successfully, as discussed in Sect.\;\ref{sec:detec}. Thus, the trace distance from the adversarial sample to the original non-adversarial state must be smaller than the distance threshold $\epsilon$. Following these rules, we have adapted the classical PGD method for the attack scenario considering the trace distance. We shall call the adapted attack TD-PGD attack. Similar to classical $l_n$ PGD attack, for each iteration, the sample is first \ding{172} perturbed up to the step size according to the gradient direction of the loss function and \ding{173} projected back to the $\epsilon$ sphere in trace distance. For our case of amplitude encoding, the trace distance threshold is equivalent to a similarity threshold on the normalized samples. We will demonstrate the method through the case of 3-dimensional input as illustrated in Fig.\;\ref{fig:TD_PGD}.

When trying to find the gradient of the model during the attack, we use a slightly different differentiable model from the training model with a softmax layer. We decreased the softmax layer temperature $t$ to $\frac{1}{50}$ to better approximate the \texttt{Argmax} function used for evaluating model accuracy. We have found that using a lower temperature does increase the effectiveness of the attack in general. However, it imposes extra overhead on the gradient calculation using a quantum computer, similar to the training overhead discussed in the \textbf{Model training} section above.

\paragraph{Bound estimation}
To estimate the bound, we follow the algorithm proposed in Sec.\;\ref{sec:alg}. The hyper-parameter we have selected for the bound estimation procedure is as follows: the number of iterations $m$ is 10. The $\alpha$ range, $[\alpha_l,\alpha_u]$, has $\alpha_l = \alpha$,  $\alpha_u = 1.1\;\alpha$, where $\alpha$ is the non-adversarial error rate given by the model in reference. The number of hyper-spheres $T$ is 20. Our hyperparameter selection for the bound estimation algorithm is far from comprehensive. However, more details on the hyper-parameter selection are shown in Appendix \ref{app:hp_selection}

\subsection{RESULTS}
In this section, we compare the experimental adversarial error with the corresponding estimated bound value for 3 instances of QVC models on MNIST and 3 instances of QCV model on FMNIST data sets specified in Sec.\;\ref{sec:setups}. We show that the models' adversarial performance after or during the training is consistently bounded by the estimated bound value.

\paragraph{Adv. error vs bound value}We produced six instances of QVC models following the setups detailed in Sec.~\ref{sec:setups}, labeled as M1, M2, M3, F1, F2, and F3. Among these, M1, M2, and M3 are quantum classifiers trained to classify the MNIST dataset, while F1, F2, and F3 are quantum classifiers for the FMNIST dataset. Three instances were selected for each dataset due to computational constraints and to demonstrate the effectiveness of the algorithm across different model configurations and datasets. We compare the adversarial error rates under the classical $l_2$ PGD attack \cite{Carlini2016TowardsNetworks} and the quantum TD-PGD attack proposed in Sec.~\ref{sec:setups}, alongside the corresponding bound values. Note that estimating the bound value requires both the non-adversarial error rate and the attack strength, leading to unique bound values for each model under different attacks. The adversarial performance of the six models on the MNIST and FMNIST datasets is summarized in Table~\ref{tab:demo_result}. It is evident from the results that the adversarial error rate consistently exceeds the estimated bound value, which indicates the consistent bounding of the adversarial error rate shown in our experiment.

\begin{table}
\centering
\begin{tabular}{| l | l | l | l | l | l|}
\multicolumn{6}{c}{\large{MNIST models}}\\
\hline
\multirow{2}{*}{Model} & Non-Adv.  & Attack & Adv. & Estimated &\multirow{2}{*}{Error $>$ Bound}\\
 & error & Str. ($\epsilon$) & error & Bound &\\
\hline
\multirow{2}{*}{M1} & \multirow{2}{*}{0.2543} & 0.1 ($l_q$) & 0.617 & 0.307 & True\\

& & 100 ($l_2$) & 0.376 & 0.273 & True\\

\multirow{2}{*}{M2}& \multirow{2}{*}{0.1601} & 0.1 ($l_q$) & 0.501 & 0.186 & True\\

& & 100 ($l_2$) & 0.243 & 0.168 & True\\

\multirow{2}{*}{M3} & \multirow{2}{*}{0.0772} & 0.1 ($l_q$) & 0.564 & 0.084 & True\\

& & 100 ($l_2$) & 0.140 & 0.082 & True\\

\hline
\multicolumn{6}{c}{}\\
\multicolumn{6}{c}{}\\
\multicolumn{6}{c}{\large{FashionMNIST models}}\\
\hline
\multirow{2}{*}{Model} & Non-Adv.  & Attack & Adv. & Estimated &\multirow{2}{*}{Error $>$ Bound}\\
& error & Str. ($\epsilon$) & error & Bound & \\
\hline
\multirow{2}{*}{F1} & \multirow{2}{*}{0.4207} & 0.1 ($l_q$) & 0.655 & 0.499 & True\\

& & 100 ($l_2$) & 0.499 & 0.442 & True\\

\multirow{2}{*}{F2} & \multirow{2}{*}{0.3179} & 0.1 ($l_q$) & 0.526 & 0.358 & True\\

& & 100 ($l_2$) & 0.362 & 0.331 & True\\

\multirow{2}{*}{F3}& \multirow{2}{*}{0.2228} & 0.1 ($l_q$) & 0.673 & 0.260 & True\\

& & 100 ($l_2$) & 0.306 & 0.237 & True\\

\hline
\end{tabular}
\caption{\textbf{
Comparison between the estimated bound value}
for 3 instances of the QVC model on the MNIST and FMNIST datasets
with different softmax temperature $t$ during training. The experimental adversarial error is always bounded by the corresponding estimated bound.
}
\label{tab:demo_result}
\end{table}

\paragraph{Performance trajectory based on training}
During the model's training, non-adversarial and adversarial errors decrease over time. To show that for any time during the training, the bond is satisfied, we have provided the trajectory in terms of non-adversarial error and adversarial error during the progress of training of 6 different models on MNIST and FashionMNIST datasets, as shown in Fig.\;\ref{fig:trajectoryC} and Fig.\;\ref{fig:trajectoryQ}. The estimated bound is the mono-color curve corresponding to the minimum possible adversarial error rate allowed by the data distribution when the given non-adversarial error rate changes. For an experiment model, it is less robust than what the data distribution allowed. From this point of view, the experimental adversarial error rate should be higher than the bound value at all times. Fig.\;\ref{fig:trajectoryC}(e)(f) and Fig.\;\ref{fig:trajectoryQ}(e)(f) have illustrated this trend as all trajectories are above the bound curve at all times.

During training these models, we used a lower learning rate for models with lower temperatures, which mitigates the higher sensitivity of the \texttt{Softmax} function regarding the quantum circuit output. All 6 models are trained with 5 epochs of 30 batches in each epoch. Thus, each batch of input contains 2000 input samples. We sample the non-adversarial and adversarial performance at batch numbers 10, 20,..., 150 to reduce the total computation time. As expected, we see a decrease in both adversarial errors and non-adversarial errors during the training. More importantly, the trajectories never cross the bound values and are also not likely to cross them with further training. This further demonstrates the reliability of using the bound value as a reference for adversarial performance evaluation during training or tunning an adversarially robust QML model.

\begin{figure*}[!htb]
    \centering
    
    \begin{subfigure}[b]{0.45\textwidth}
        \centering
        \includegraphics[width=\textwidth]{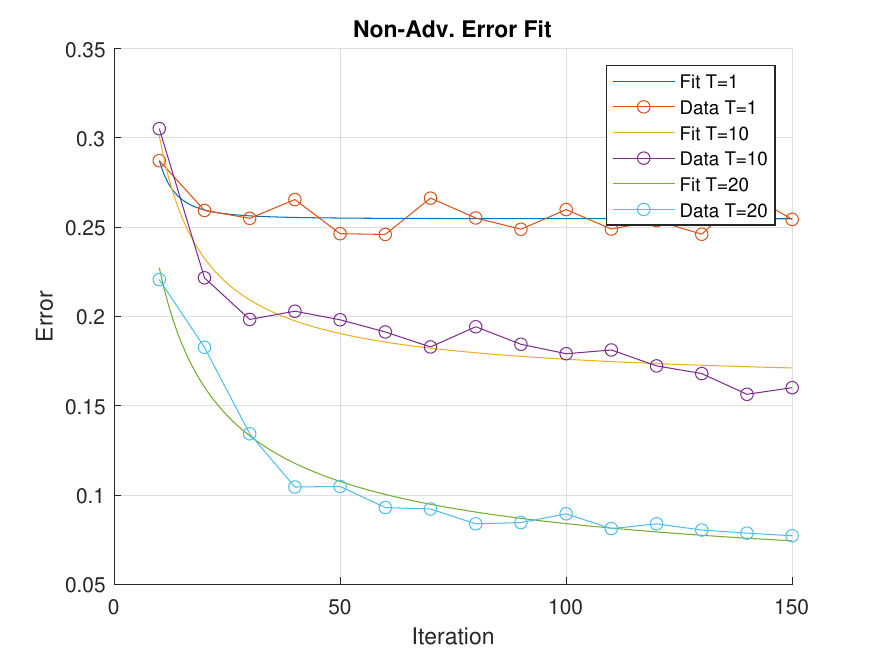}
        \caption{MNIST Non-Adv. Error vs Iteration}
    \end{subfigure}
    \hfill
    \begin{subfigure}[b]{0.45\textwidth}
        \centering
        \includegraphics[width=\textwidth]{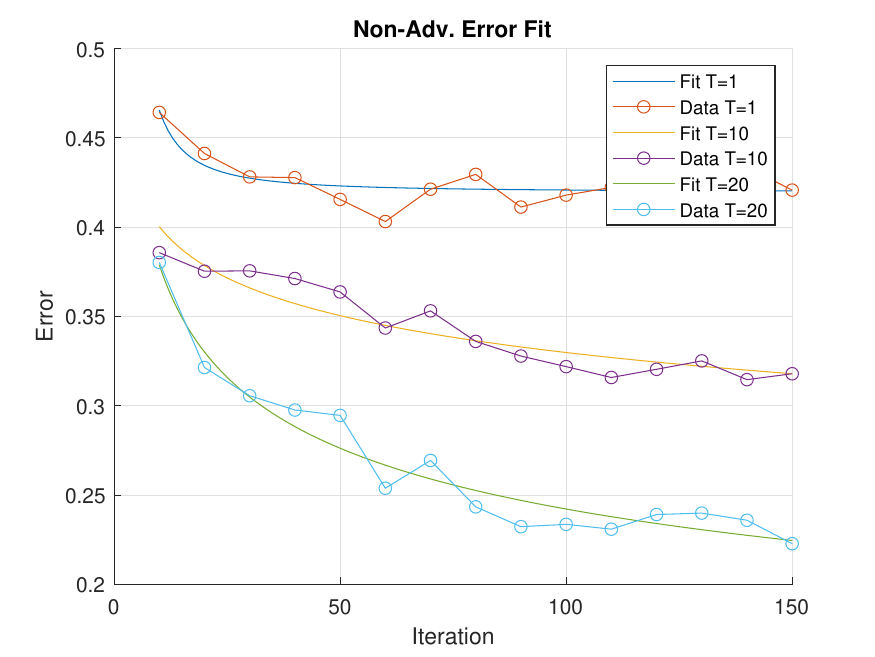}
        \caption{FMNIST Non-Adv. Error vs Iteration}
    \end{subfigure}

    \begin{subfigure}[b]{0.45\textwidth}
        \centering
        \includegraphics[width=\textwidth]{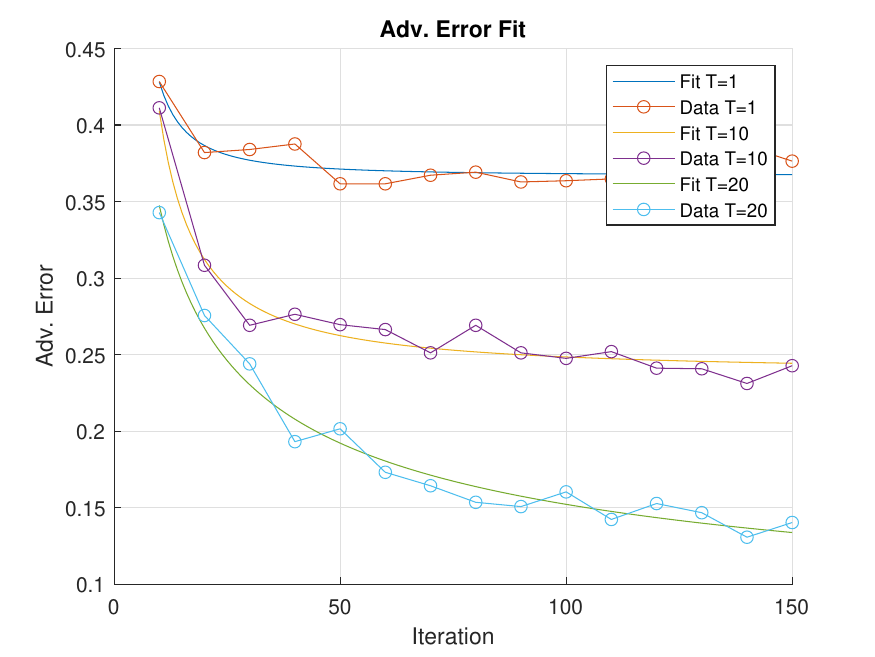}
        \caption{MNIST Adv. Error vs Iteration}
    \end{subfigure}
    \hfill
    \begin{subfigure}[b]{0.45\textwidth}
        \centering
        \includegraphics[width=\textwidth]{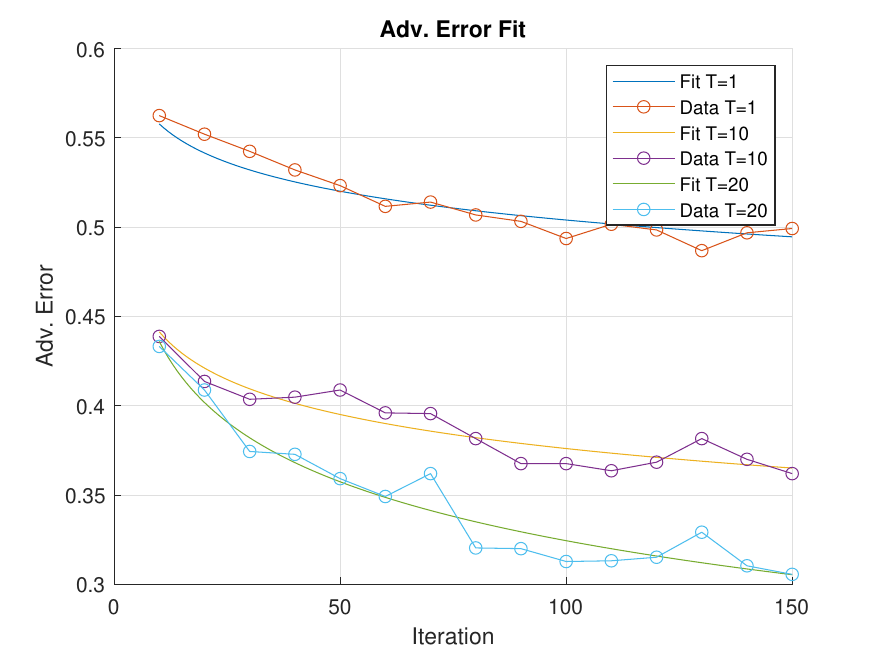}
        \caption{FMNIST Adv. Error vs Iteration}
    \end{subfigure}
    
    \caption{\textbf{The evolution of non-adversarial and adversarial error rates under classical perturbation attacks}. The results show the error rate evolution during the training of 3 different instances of QVC models on MNIST and FMNIST datasets. The attack is $l$-PGD attack of strength 
    $\epsilon=100/255$. (a) and (b) shows the estimates of the non-adversarial error rate. Each data point is generated by 10000 samples in the testing sets of MNIST and FMNIST datasets. (b) and (c) shows the estimates of the adversarial error rate. Each data point is generated by 2500 samples of 250 samples per class. This ensures that the standard deviation of the adversarial error rate is smaller than 0.04. For the aforementioned 4 diagrams, we also least-square fitted power functions with negative power to the estimated error. These results are used to construct Fig.~\ref{fig:trajectoryC}.
    } 

    \label{fig:historyC}
\end{figure*}

\begin{figure*}[!htb]
    \centering
    
    \begin{subfigure}[b]{0.45\textwidth}
        \centering
        \includegraphics[width=\textwidth]{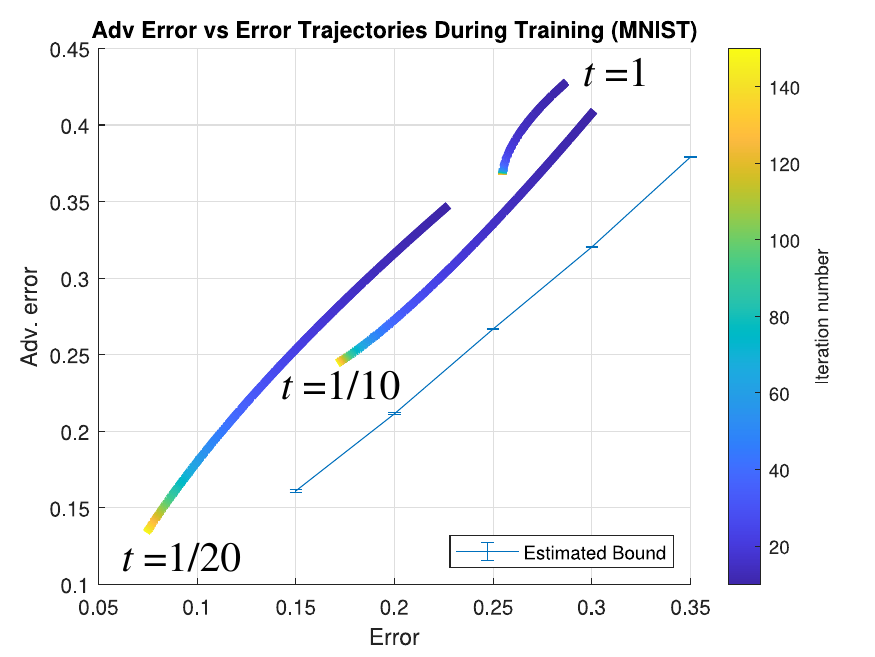}
        \caption{MNIST Trajectory}
    \end{subfigure}
    \hfill
    \begin{subfigure}[b]{0.45\textwidth}
        \centering
        \includegraphics[width=\textwidth]{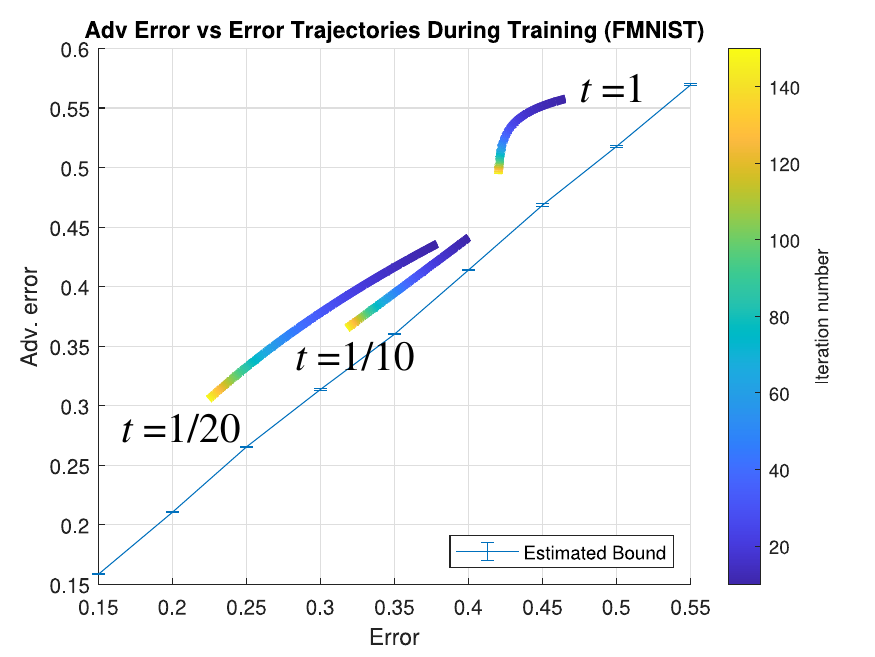}
        \caption{FMNIST Trajectory}
    \end{subfigure}
    
    \caption{\textbf{The trajectory of performances under classical $l_2$ attack.} (a) and (b) shows the evolution of the model in terms of adversarial and non-adversarial error during training based on the least-squared fitted curve in Fig.\;\ref{fig:historyC}. The trajectory is compared with the estimated bound at different levels of non-adversarial error with the training adversarial errors. This shows that the estimated bound always lower bound the adversarial error during training in the 6 instances of QVC models.}

    \label{fig:trajectoryC}
\end{figure*}

\begin{figure*}[!htb]
    \centering
    
    \begin{subfigure}[b]{0.45\textwidth}
        \centering
        \includegraphics[width=\textwidth]{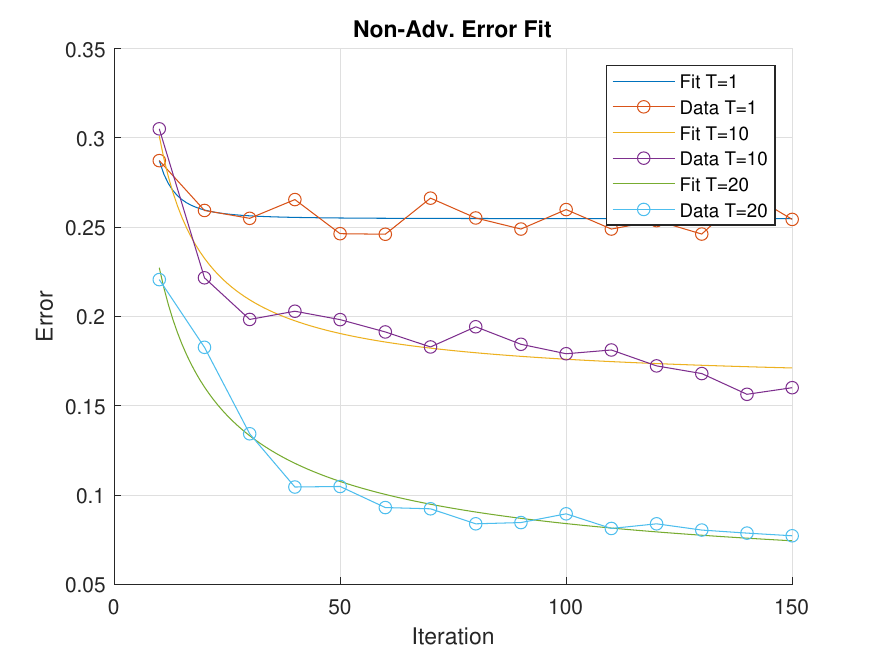}
        \caption{MNIST Non-Adv. Error vs Iteration}
    \end{subfigure}
    \hfill
    \begin{subfigure}[b]{0.45\textwidth}
        \centering
        \includegraphics[width=\textwidth]{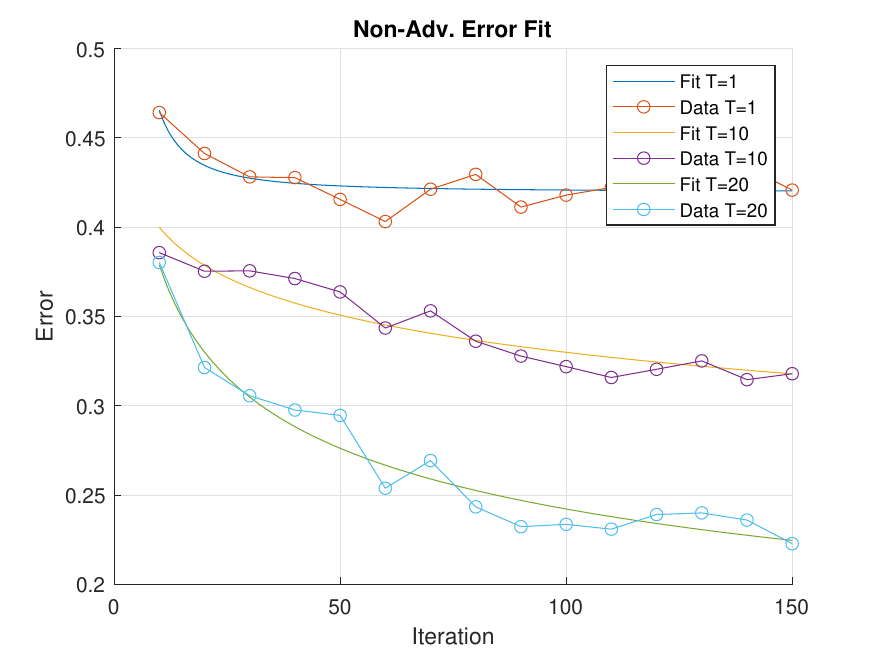}
        \caption{MNIST Non-Adv. Error vs Iteration}
    \end{subfigure}

    \begin{subfigure}[b]{0.45\textwidth}
        \centering
        \includegraphics[width=\textwidth]{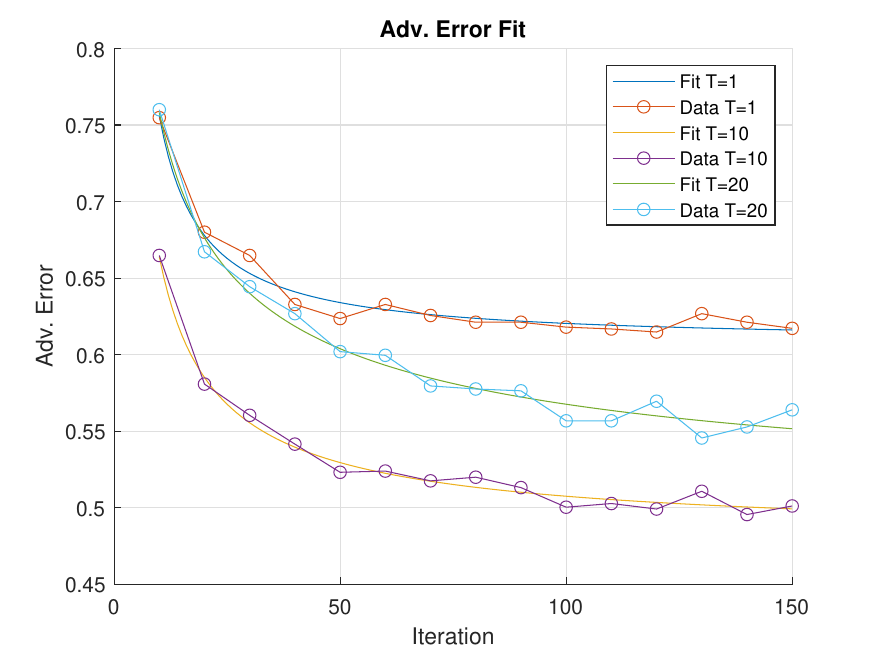}
        \caption{MNIST Adv. Error vs Iteration}
    \end{subfigure}
    \hfill
    \begin{subfigure}[b]{0.45\textwidth}
        \centering
        \includegraphics[width=\textwidth]{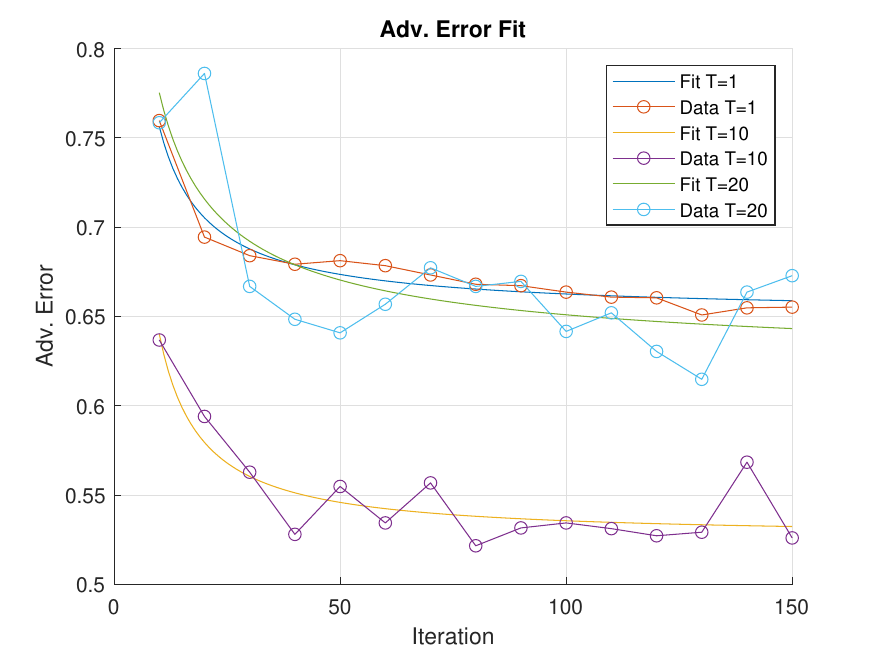}
        \caption{FMNIST Adv. Error vs Iteration}
    \end{subfigure}
    
    \caption{\textbf{The evolution of non-adversarial and adversarial error rates during training} for 6 instances of the QVC model on MNIST and FMNIST datasets. The attack is a TD-PGD attack of strength $\epsilon=0.1$. Following the presentation of Fig.\;\ref{fig:trajectoryC}, (a) (b) (c) (d) shows the non-adversarial and adversarial evolution with iteration. These results are used to construct Fig.\;\ref{fig:trajectoryQ}.}

    \label{fig:historyQ}
\end{figure*}

\begin{figure*}[!htb]
    \centering
    
    \begin{subfigure}[b]{0.45\textwidth}
        \centering
        \includegraphics[width=\textwidth]{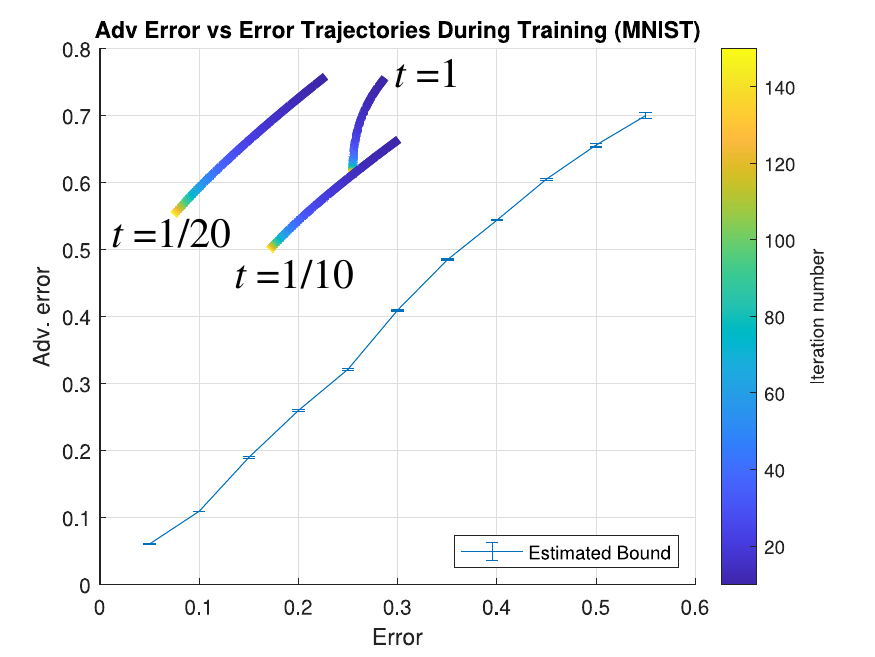}
        \caption{MNIST Trajectory}
    \end{subfigure}
    \hfill
    \begin{subfigure}[b]{0.45\textwidth}
        \centering
        \includegraphics[width=\textwidth]{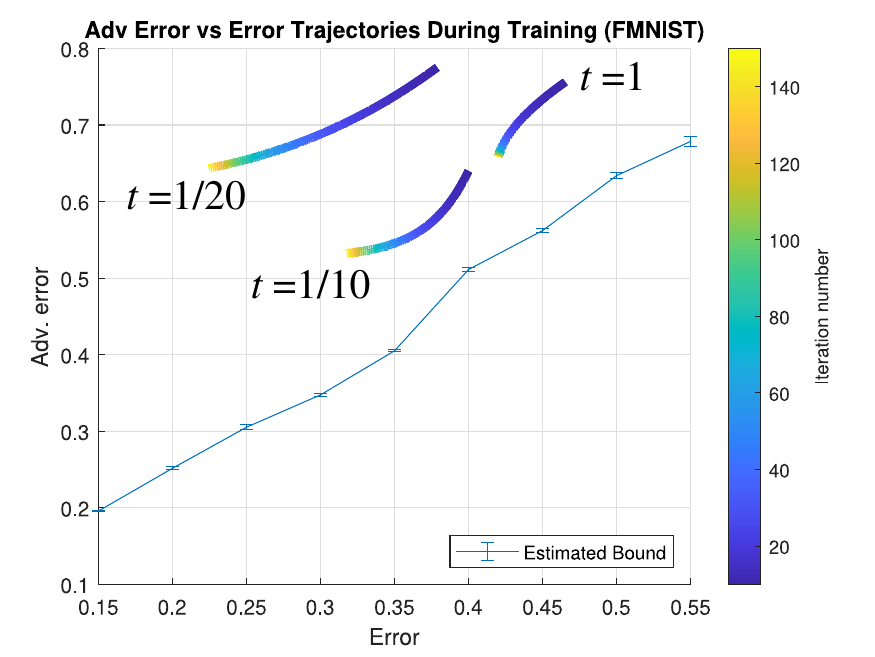}
        \caption{FMNIST Trajectory}
    \end{subfigure}
    
    \caption{\textbf{The trajectory of performances under quantum perturbation attack.} (a) and (b) shows the evolution of the model in terms of adversarial and non-adversarial error during training based on the least-squared fitted curve in Fig.\;\ref{fig:historyQ}. The trajectory is compared with the estimated bound at different levels of non-adversarial error with the training adversarial errors. This shows that the estimated bound always lower bound the adversarial error during training in the 6 instances of QVC models.}

    \label{fig:trajectoryQ}
\end{figure*}

\paragraph{Selection of Softmax temperature parameter}

\begin{table*}
\centering
\begin{tabular}{|c|c|}
\hline
\begin{tabular}[c]{@{}c@{}}
\begin{tabular}{|c|c|c|c|}
\multicolumn{4}{c}{MNIST $l_2$ attack}\\
\hline
\multirow{2}{*}{Model} & Softmax & Non-Adv. & \multirow{2}{*}{$\frac{\textrm{Adv. error}}{\textrm{Bound}}$} \\ 
 & temp. ($t$) & error & \\ \hline
M1 & 1 & 0.2543 & 1.37 \\ \hline
M2 & $\frac{1}{10}$ & 0.1601 & 1.45 \\ \hline
M3 & $\frac{1}{20}$ & 0.0772 & 1.71 \\ \hline
\end{tabular}
\end{tabular}
& 
\begin{tabular}[c]{@{}c@{}}
\begin{tabular}{|c|c|c|c|}
\multicolumn{4}{c}{MNIST $l_q$ attack}\\
\hline
\multirow{2}{*}{Model} & Softmax & Non-Adv. & \multirow{2}{*}{$\frac{\textrm{Adv. error}}{\textrm{Bound}}$} \\ 
 & temp. ($t$) & error & \\ \hline
M1 & 1 & 0.2543 & 2.01 \\ \hline
M2 & $\frac{1}{10}$ & 0.1601 & 2.69 \\ \hline
M3 & $\frac{1}{20}$ & 0.0772 & 6.71 \\ \hline
\end{tabular}
\end{tabular}
\\ \hline
\begin{tabular}[c]{@{}c@{}}
\begin{tabular}{|c|c|c|c|}
\multicolumn{4}{c}{FMNIST $l_2$ attack}\\
\hline
\multirow{2}{*}{Model} & Softmax & Non-Adv. & \multirow{2}{*}{$\frac{\textrm{Adv. error}}{\textrm{Bound}}$} \\ 
 & temp. ($t$) & error & \\ \hline
F1 & 1 & 0.4207 & 1.13 \\ \hline
F2 & $\frac{1}{10}$ & 0.3179 & 1.09 \\ \hline
F3 & $\frac{1}{20}$ & 0.2228 & 1.29 \\ \hline
\end{tabular}
\end{tabular}
& 
\begin{tabular}[c]{@{}c@{}}
\begin{tabular}{|c|c|c|c|}
\multicolumn{4}{c}{FMNIST $l_q$ attack}\\
\hline
\multirow{2}{*}{Model} & Softmax & Non-Adv. & \multirow{2}{*}{$\frac{\textrm{Adv. error}}{\textrm{Bound}}$} \\ 
 & temp. ($t$) & error & \\ \hline
F1 & 1 & 0.4207 & 1.31 \\ \hline
F2 & $\frac{1}{10}$ & 0.3179 & 1.47 \\ \hline
F3 & $\frac{1}{20}$ & 0.2228 & 2.59 \\ \hline
\end{tabular}
\end{tabular}
\\ \hline
\end{tabular}
\caption{\textbf{The effects of softmax temperature $t$ on non-adversarial error and general adversarial performance.} Instead of the adversarial error rate shown in Table \ref{tab:demo_result}, we use the comparative value of Adv. error/Bound for indication of robustness. For the same attack, we regard models with a lower ratio as closer to the optimal model suggested by the bound value. Generally, a higher softmax temperature in training leads to a lower ratio.}
\label{tab:bound_compare}
\end{table*}

The 6 instances of QVC models for MNIST and FMNIST datasets are trained with different softmax temperature $t$. Among these 6 models, M1 and F1 have the highest temperature at $t=1$ while M3 and F3 have the lowest temperature at $t=1/20$. With a lower temperature, the \texttt{Softmax} function is sharper and more closely resembles the \texttt{Argmax} function. This sharpness of the softmax layer at the end of the model pipeline results in a lower non-adversarial error rate, as seen in Table \ref{tab:demo_result}. However, this also impacts the adversarial performance, as Table \ref{tab:bound_compare} highlighted. In Table \ref{tab:bound_compare}, we use the ratio between adversarial error and the estimated bound value to indicate how close the model is to the optimal robust model. The results show that with a decrease of $t$, the ratio tends to increase, suggesting an increase in robustness. The outlier to this trend is F2 under $l_2$ attack, which shows a ratio slightly lower than that of F1. The literature on the effects of softmax temperature for the QML model is limited, and more substantial research is required to verify the trend and understand the outlier of this trend.

\paragraph{Discussion}
The experiment results above show the validity of the adversarial risk lower bound by evaluating the performances of several quantum models on the MNIST and FMNIST datasets. However, the usefulness of any bound also depends on how tight it is or, in other words, how achievable it is in practice. For the 6 quantum models we have trained, we see a moderate amount of adversarial error rate deviation from the bound. The deviation is on the same order of magnitude as the bound and sometimes less than 10\% in the best case. Other robust machine learning techniques, such as adversarial training and regularization, may help to approach the bound. Nonetheless, approaching the bound with near-optimal models remains an open question in both CML and QML.

The adversarial perturbation generated by these attacks exhibits systematic features that align with human intuition to some extent, particularly for models trained with higher softmax temperatures. In Table \ref{tab:examples}, we see the attack aims to alter the true label of the image. For the MNIST dataset, the attack adds horizontal strokes to sample 0 and thickens the stroke of sample 1. For the FMNIST dataset, the attack adds sleeves to the T-shirt and removes sleeves for the coat. This alignment for quantum models has been compared with similar behavior for robust and non-robust classical models \cite{West2023BenchmarkingScale}. 

This phenomenon has been studied in the classical literature and is regarded as typical behavior of models designed to be robust \cite{Ilyas2019AdversarialFeatures, Tsipras2018RobustnessAccuracy}. As highlighted in \cite{West2023BenchmarkingScale}, this seemingly intrinsic robustness of the quantum model without any adversarial training is an interesting topic. Nonetheless, for QML, our comparison between quantum models with different accuracies highlights the concern of degrading adversarial sample interpretability and, thus, model robustness when the model accuracy increases. 

By examining the ratio between the model-specific adversarial error and the bound value for the 6 instances of models, we observe a decreasing trend in the ratio when the clean learning error rate decreases. It indicates the relative robustness (compared to the bound) and accuracy trade-off but in a higher error regime. The classic trade-off between (absolute) robustness and accuracy has been formalized in \cite{Tsipras2018RobustnessAccuracy}, where the inherent uncertainty of labeling in the data distribution often emerges to impose a trade-off between the non-adversarial and adversarial performance. We have discussed the adversarial risk lower bound and its connection with experimental adversarial error in Sec.\;\ref{sec:bound}. The direct reference value of the bound for optimizing robust models is facilitated by the assumption of robust ground truth. However, the uncertainty in original data labeling may emerge in certain high levels of non-adversarial accuracy when the perturbation strength is also high enough. This uncertainty will violate the assumption we have set up and further increase the adversarial error rate, making the bound less tight. In these circumstances, we may rely on other theoretical or experimental bounds that could be tighter than the estimated bound we proposed.

\begin{table*}
\centering 

\begin{minipage}{0.45\linewidth}
\centering
\begin{tabular}{m{2cm} m{0.1\linewidth} m{0.1\linewidth} m{0.1\linewidth} m{0.1\linewidth}}

\multicolumn{5}{c}{\huge{MNIST models}}\\
\hline
 & \multicolumn{2}{c}{\large{$l_2$ attack}} & \multicolumn{2}{c}{\large{$l_q$ attack}}\\

\textbf{Clean Samples} & \includegraphics[width=\linewidth]{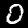} & \includegraphics[width=\linewidth]{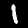} 
& \includegraphics[width=\linewidth]{pic/Clean/test_m1_72.png} & \includegraphics[width=\linewidth]{pic/Clean/test_m1_75.png}\\

\textbf{M1} & \includegraphics[width=\linewidth]{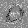} & \includegraphics[width=\linewidth]{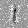}
& \includegraphics[width=\linewidth]{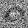} & \includegraphics[width=\linewidth]{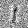}\\

\textbf{M2} & \includegraphics[width=\linewidth]{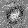} & \includegraphics[width=\linewidth]{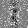}
& \includegraphics[width=\linewidth]{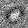} & \includegraphics[width=\linewidth]{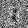}\\

\textbf{M3} & \includegraphics[width=\linewidth]{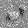} & \includegraphics[width=\linewidth]{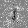}
& \includegraphics[width=\linewidth]{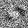} & \includegraphics[width=\linewidth]{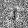}\\

\end{tabular}
\end{minipage}%
\hfill
\begin{minipage}{0.45\linewidth}
\centering
\begin{tabular}{m{2cm} m{0.1\linewidth} m{0.1\linewidth} m{0.1\linewidth} m{0.1\linewidth}}

\multicolumn{5}{c}{\huge{FMNIST models}}\\
\hline
 & \multicolumn{2}{c}{\large{$l_2$ attack}} & \multicolumn{2}{c}{\large{$l_q$ attack}}\\

\textbf{Clean Samples} & \includegraphics[width=\linewidth]{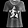} & \includegraphics[width=\linewidth]{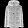} 
& \includegraphics[width=\linewidth]{pic/Clean/ctest_f1_6.png} & \includegraphics[width=\linewidth]{pic/Clean/ctest_f1_8.png}\\

\textbf{F1} & \includegraphics[width=\linewidth]{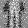} & \includegraphics[width=\linewidth]{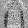}
& \includegraphics[width=\linewidth]{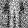} & \includegraphics[width=\linewidth]{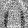}\\

\textbf{F2} & \includegraphics[width=\linewidth]{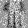} & \includegraphics[width=\linewidth]{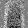}
& \includegraphics[width=\linewidth]{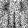} & \includegraphics[width=\linewidth]{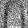}\\

\textbf{F3} & \includegraphics[width=\linewidth]{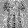} & \includegraphics[width=\linewidth]{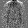}
& \includegraphics[width=\linewidth]{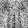} & \includegraphics[width=\linewidth]{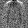}\\

\end{tabular}
\end{minipage}
\caption{\textbf{The adversarial perturbation generated by the attacks.} The attack with attack strength $\epsilon=100/255\;(l_2)$ and $\epsilon=0.1\;(l_q)$ are applied on the 6 instances of QVC models on MNIST and FMNIST datasets. In general, models with a higher training softmax temperature $t$ (M1 and F1) exhibit a greater extent of systematic features. The systematic features capture the intuitive steps towards changing the labels of the image (i.e., Adding horizontal lines to sample 0, thickening the stroke of sample 1 for the MNIST dataset; For the FMNIST dataset, adding sleeves to the T-shirt, removing sleeves from the coat). As the higher accuracy model (M3, F3) exhibits less systematic features, the interpretability and, thus, the robustness of the model degrades.}
\label{tab:examples} 
\end{table*}

\section{CONCLUSION}

In this work, we establish a connection between the experimental adversarial error rate and the theoretical adversarial risk lower bound in the quantum machine learning problem, justifying the use of the latter as a benchmark when evaluating the resilience of non-optimal models against potent attacks. To assess the algorithm's effectiveness in evaluating adversarial performance, we attack some quantum models with quantum perturbation attack strategies, including an attack we devised for the quantum perturbation scenario. Then, we compare the bound with the adversarial error rates for the simulated quantum models. We observed the bounding of the error rates to different extents depending on the hyper-parameters chosen, indicating the utility of the bound to identify robust models. The lower bound represents the minimal adversarial error achievable through model optimization against a robust attack, serving as an objective for both machine learning practitioners and a performance estimate in general. 


In the future, incremental experimental studies on the adversarial risk lower bound estimation can further explore other datasets and quantum encoding techniques, such as parametric-gate-based encoding. Further algorithmic studies may also extend the usage or optimize the efficiency of the bound calculation with quantum computers. Utilizing the bound, studies on the robust QML architecture and circuit design can anchor their results based on the theoretically allowed robustness for the given error rate instead of comparing it with a potentially very different classical model in terms of learning error rates. These efforts help bring models closer to, or even achieve, the robustness limit, i.e., the optimal defenses against adversarial attacks.

\appendices


\section{PROOF OF TRIANGLE INEQUALITY OF ANGLES BETWEEN COMPLEX UNIT VECTORS}\label{app:tri}
The Bures angles between the two arbitrary complex unit vectors are defined as  $\theta_{w,u} \equiv \arccos |w^\dagger \cdot u|$. In the context of quantum states, equivalently, we have $\theta_{w,u} \equiv \arccos \braket{w}{u}$.

Using Bures angle, we have the following theorem:
\begin{theorem}[triangle inequality for Bures angles]
Given 3 arbitrary complex unit vector $v_1$, $v_2$ and $v_3$ in any dimension $d$, their Bures angle between each other satisfy:
\begin{equation}
\end{equation}
\end{theorem}

To prove the theorem above, consider the Gram matrix $G_{ij} = \braket{v_i}{v_j}$:
\begin{equation}
    G =
    \begin{bmatrix}
    1 & \braket{v_1}{v_2} & \braket{v_1}{v_3}\\
    \braket{v_2}{v_1} & 1 & \braket{v_2}{v_3}\\
    \braket{v_3}{v_1} & \braket{v_3}{v_2} & 1
    \end{bmatrix}\;.
\end{equation}
The determinant of the Gram matrix $G$ is
\begin{multline}
    \det(G) = 1-\cos^2(\theta_{v_1,v_2})-\cos^2(\theta_{v_2,v_3})-\cos^2(\theta_{v_1,v_3}) + 
    \\
    2\textrm{Re}[\braket{v_1}{v_2} \braket{v_2}{v_3} \braket{v_3}{v_1}]\;.
\end{multline}
By the definition of the Bures angle, $|\braket{v_1}{v_2} \braket{v_2}{v_3} \braket{v_3}{v_1}| = \cos(\theta_{v_1,v_2})\cos(\theta_{v_2,v_3})\cos(\theta_{v_1,v_3})$. Thus,
\begin{multline}
\label{eq:c1}
    \det(G) \leq 1-\cos^2(\theta_{v_1,v_2})-\cos^2(\theta_{v_2,v_3})-\cos^2(\theta_{v_1,v_3}) + 
    \\
    2\cos(\theta_{v_1,v_2})\cos(\theta_{v_2,v_3})\cos(\theta_{v_1,v_3})\;.
\end{multline}
Combine this with the fact of the determinant of the Gram matrix being non-negative, 
\begin{equation}
\label{eq:GramDet}
    \det(G) \geq 0
\end{equation}
we have
\begin{multline}
    1-\cos^2(\theta_{v_1,v_2})-\cos^2(\theta_{v_2,v_3})-\cos^2(\theta_{v_1,v_3}) + \\2 \cos(\theta_{v_1,v_2})\cos(\theta_{v_2,v_3})\cos(\theta_{v_1,v_3}) \geq 0 \;.
\end{multline}
Simplify the inequality with all angles within $[0,\;\pi]$,
\begin{equation}
    \cos(\theta_{v_1,v_3}) \geq \cos(\theta_{v_1,v_2} + \theta_{v_2,v_3})\;.
\end{equation}
This implies
\begin{equation}
    \theta_{v_1,v_2}+\theta_{v_2,v_3}\geq\theta_{v_1,v_3}\;.
\end{equation}

The equality sign of (\ref{eq:c1}) is achieved if and only if $\braket{v_1}{v_2} \braket{v_2}{v_3} \braket{v_3}{v_1}$ is a real number. In addition, the equality sign of (\ref{eq:GramDet}) is reached if and only if the vectors $v_1, v_2, v_3$ are linearly dependent. Thus, the triangle inequality of angles achieves equality if and only if both of the two conditions above are satisfied. 

\section{The Bound Estimation Algorithms}\label{app:alg}
This section details our algorithms' pseudocode for finding robust error regions for either classical $l_2$ distance or quantum trace distance. We also provide the runtime analysis and compare it with the runtime of a similar algorithm proposed for the classical attack scenario \cite{MahloujifarEmpiricallyRobustness}. 

\subsection{PSEUDOCODE}
The pseudocode is shown in Algorithm \ref{alg:loops},  Algorithm \ref{alg:Condense} and Algorithm \ref{alg:Bound_Estimation}.  Algorithm \ref{alg:loops} is the core subroutine in Algorithm \ref{alg:Bound_Estimation}. The Algorithm \ref{alg:Condense} shows the construction of the crucial function \texttt{Condense} in Algorithm \ref{alg:loops} that enable the parallelised computation of the bound.

Algorithm \ref{alg:loops} fundamentally follows a heuristic greedy optimization approach. It pre-computes and tracks the pair-wise distances between samples when more and more samples are taken into the error regions. After carving out multiple hyper-spheres from the input data space, we are left with input data that are not included in the hyper-spheres, i.e., $S \setminus S_\mathcal{E}$ and $S \setminus S'_\mathcal{E}$ shown in Algorithm \ref{alg:loops}. They represent the non-adversarial and adversarial error rate at this stage. To efficiently optimize the centers and radii for the next iteration utilizing the sorted pair-wide distance $D^{(s)}$, we condense$D^{(s)}$ to $D^{(s,c)}$, i.e., the distances between all $N$ samples (potential centers of hyper-sphere) and the remaining points in $S \setminus S_\mathcal{E}$ and $S \setminus S'_\mathcal{E}$. In the Algorithm~\ref{alg:loops}, we denote this step as the \texttt{Condense} function. We provide a pseudo-code that demonstrates the \texttt{Condense} function in Algorithm~\ref{alg:Condense}.

The \texttt{FindRoot} function in Algorithm~\ref{alg:loops} is a root-finding function utilizing the bisection method to find the element closest to zero in an ascending list. This method is deterministic with a complexity of $O(\log N_{\textrm{list}})$, where $N_\textrm{list}$ is the number of elements in the ascending list. The \texttt{Expand} function in Algorithm~\ref{alg:loops} is $r'=r+\epsilon$ for classic $l_n$ distances and $r'=r\sqrt{1-\epsilon^2} + \epsilon\sqrt{1-r^2}$ for the quantum trace distance, as discussed in Sect.\;\ref{sec:Qadapt}.

Algorithm \ref{alg:Condense} outlines the \texttt{Condense} function, which consists of two steps. The first step is to compute the unsorted to sorted location index from the index output $I$ of a common sorting algorithm such as Quicksort. The matrix $I'$ shows where each element in the corresponding location of $D$ is moved to $D^{(s)}$. The second step is to remove all the elements in $D^{(s)}$ that correspond to distances from any sample in $S$ (indexed by $i$) to samples in $S_\mathcal{E}$ (indexed by $k$). Using the location index $I'$, we can efficiently locate them and remove them from $D^{(s)}$. If necessary, each of the two steps above can be programmed to run on GPU with $(i,j)$ or $(i,k)$ parallel separately.

\SetKwComment{Comment}{/* }{ */}
\SetKwFor{For}{for (}{) $\lbrace$}{$\rbrace$}
\begin{algorithm*}
\caption{\texttt{Heuristic Search}: Paralleled Heuristic Search for Robust Error Region}\label{alg:loops}
\KwData{Pair-wise distances $D(x_i,x_j)$ between arbitrary $i$th and $j$th samples in the sample set $S$ under the given metric; perturbation strength $\epsilon$; error threshold $\alpha$; number of hyper-spheres $T$}
\KwResult{The error region $\mathcal{E}$ defined by the set of centres $C=\{x_i\}$ and the set of corresponding radius $R=\{r_i\}$}

$D^{(s)}$, $I$ = Sort($D$, axis=1) \Comment*[r]{$D^{(s)}$ is the sorted pair-wise distances}
\Comment*[r]{$I$ is the corresponding original $j$ index of values in $D^{(s)}_{ij}$}

$S_\mathcal{E} \gets [~],~ S_\mathcal{E}' \gets [~]$ \Comment*[r]{$S_\mathcal{E}'$ is the perturbed error set}
$C \gets [~],~ R \gets [~]$\;
$D^{(s,c)} = D^{(s)}$\;
$D'^{(s,c)} = D^{(s)}$\;
\For{$t=0;\; t<T;\;t=t+1$}{
    $k_l = \lceil{(\alpha \abs{S} - \abs{S_{\mathcal{E}}})/(T-t)}\rceil$,\;
    $k_u=\lceil(\alpha \abs{S} - \abs{S_{\mathcal{E}}})\rceil$\; \Comment*[r]{Choose a reasonable range for the number of point included for this sphere}
    \For{$k=k_l;\;k\leq k_u;\;k=k+1$}{
        \For{$i=0;\;i\leq\abs{S};\;i=i+1$}{
            $r_{ik}=D^{(s,c)}_{ik}$ \Comment*[r]{The radius of sphere at $x_i$ including $k$ samples }
            $r'_{ik}=\texttt{expand}(r_{ik},\;\epsilon)$ \Comment*[r]{For $l_n$, $r' = r + \epsilon$. For $l_q$, $r'=r\sqrt{1-\epsilon^2} + \epsilon\sqrt{1-r^2}$}
            $k'= \texttt{FindRoot}_k(D'^{(s,c)}_i-r'_{ik})$ \Comment*[r]{error sample number under adversary}
            $\Delta_{ik} = k'-k$ \Comment*[r]{The increase of error sample}
            }
    }
    $(i, k) = \texttt{Argmin}(\Delta_{ik})$\;
    $S_\mathcal{E} = S_\mathcal{E}\cup S_{ik}$ \Comment*[r]{$S_{ik}$ is the subset of samples within $Sphere(x_i,\;r_{ik})$}
    $S'_\mathcal{E} = S'_\mathcal{E}\cup S'_{ik}$ \Comment*[r]{$S'_{ik}$ is the subset of samples within $Sphere(x_i,\;r'_{ik})$}
    $D^{(s,c)}$\texttt{ = Condense}($D^{(s)}$,  $S_\mathcal{E}$,  $I$) \Comment*[r]{Condense to all distances concerning $S \setminus S_\mathcal{E}$}
    $D'^{(s,c)}$\texttt{ = Condense}($D^{(s)}$,  $S'_\mathcal{E}$,  $I$) \Comment*[r]{Condense to all distances concerning $S \setminus S'_\mathcal{E}$}
    Append $x_i$ to C\;
    Append $r_{ik}$ to R\;
}
$\mathcal{E} = \bigcup_{i=0}^T Sphere(C_i, R_i)$
\end{algorithm*}

\begin{algorithm*}
\caption{\texttt{Condense}: Sorted Pair-wise Distance Condense \label{alg:Condense}}
\KwData{Sorted pair-wise distances $D^{(s)}$; The set of samples already in the error region $S_\mathcal{E}$; The order of the sorted pair-wise distances $I$}
\KwResult{The condensed pair-wise distances $D^{(s,c)}$ from all samples to samples in $S \setminus S_\mathcal{E}$}
\For{$i=0;\;i<N;\;i=i+1$}{
        \For{$j=0;\;j<N;\;j=j+1$}{
        $I'_{i,I_{ij}} = j$ \Comment*[r]{Construct the unsorted to sorted location index}
        }
}
\Comment*[r]{The matrix $I'$ should be at the same size as $I$ and completely filled out}

$D^{(s,c)}=D^{(s)}$;

\For{$i=0;\;i<N;\;i=i+1$}{
        \For{$\forall k\;|\;x_k \in S_\mathcal{E}$}{
        $\texttt{Remove Elements of the Set at Index} (\texttt{Set}=D_i^{(s,c)}, \texttt{Index}=I'_{ik})$;
        }
}
\Comment*[r]{$D^{(s,c)}$ should be a rectangular array with size ($\#S$, $\#(S \setminus S'_\mathcal{E})$)}
\end{algorithm*}

\begin{algorithm*}
\caption{Bound Estimation Using Robust Error Region Search \label{alg:Bound_Estimation}}
\KwData{The sample set $S = \{x_i\}$; a pair-wise distance metric, $l(\cdot, \cdot)$; perturbation strength $\epsilon$; error threshold $\alpha$; number of hyper-spheres $T$, number of iteration, $m$; Search $\alpha$ range, $[\alpha_l, \alpha_u]$}
\KwResult{The estimated bound value, $c_{\textrm{adv}}$}

\For{$ite=0;\;ite<m;\;ite=ite+1$}{
        $\alpha_{\textrm{step}} = (\alpha_u - \alpha_l) / m$\;
        $\alpha_{\nu} = \alpha_l + ite\;\alpha_{\textrm{step}}$\;
        $S_\textrm{train}, S_\textrm{test} = \texttt{Random Partition}(S)$\;
        $\forall x_i,\;x_j \in S_\textrm{train} \;\; d_{ij} = l(x_i, x_j)$ \Comment*[r]{Compute the pairwise distance for the training}
        $\mathcal{E} = \texttt{} \texttt{Heuristic Search}(d,\;\epsilon,\;\alpha_{\nu},\;T)$\;
        $Risk_{\nu} = \#\{x_i\;|\;x_i\in S_{\textrm{test}},\;x_i\in\mathcal{E}\}\; / \;\#S$\;
        $AdvRisk_{\nu} = \#\{x_i\;|\;x_i\in S_{\textrm{test}},\;x_i\in\mathcal{E}_\epsilon\} \;/\; \#S$\;
}
$\textrm{RegressionModel}(Risk) = \texttt{Regression}(Risk_{\nu},\;AdvRisk_{\nu})$\;
$c_\textrm{adv} = \textrm{RegressionModel}(\alpha)$

\end{algorithm*}

\subsection{RUNTIME ANALYSIS}
In this section, we analyze the complexity of the algorithm and focus on the scaling behaviour of the runtime regarding the data dimension $d$ and sample size $n$. Thus, We deliberately omit all hyper-parameter constants such as $T$ and $\alpha$ in the discussion of complexity and focus on the data dimension and sample number instead. However, we note that for a similar algorithm in \cite{MahloujifarEmpiricallyRobustness}, the sample size $T$ need to increase at $O(n^{1/4})$ so that the bound estimate converges to the adversarial risk when $n\rightarrow \infty$.

For Algorithm \ref{alg:loops}, the proposed method allocates the calculation of distances between samples and the k-nearest neighbor to the preprocessing step before any optimization procedures. The preprocessing step has complexity $O(n^2 d)$. For the main loops of choosing hyper-spheres, the inner loop indexed by $k$ and $i$ can be executed on GPU in parallel without any approximation. The \texttt{Condense} function has complexity at $O(n^2)$, which is insignificant compared to the more costly part, the \texttt{FindRoot} function in the loops of Algorithm~\ref{alg:loops}. Thus, dominated by the \texttt{FindRoot} function with complexity $O(\log n)$ (each call), the main loops have a complexity of $O(n^2 \log n)$. When considering the regression regarding testing Adversarial risk and Non-adversarial risk, an additional multiplicative constant factor will be introduced, which will not affect the overall complexity.

The $l_2$ approach in \cite{MahloujifarEmpiricallyRobustness} utilized Ball Tree to preprocess and store the k nearest neighbors for each iteration. This results in a lower complexity for the preprocessing step at $O(nd \log n)$ but higher complexity in main loops of $O(n^2 d)$. Compare this with the complexity given by our algorithm; the difference factor is $d/\log n$. Assuming the logarithmic complexity regarding the sample number $O(\log n)$ poses less challenge than the complexity of data dimension $d$, our algorithms will be faster in runtime complexity. Additionally, the Ball Tree algorithm is generally unsuitable to run on GPU architecture, which is a more prevalent and scalable computing architecture at the moment. In addition, when estimating the bound for the quantum perturbation attack we have proposed, the pair-wise trace distance can be inferred from a swap test \cite{Buhrman2001QuantumFingerprinting} or an overlap estimation algorithm \cite{Cincio2018LearningQuantumOverlap, Fanizza2020BeyondSwapTest}. These quantum-based distance estimation algorithms will significantly decrease the time complexity of distance calculation, making the time complexity even more dominated by the main loops of the Algorithm \ref{alg:loops}.

\section{Hyper-parameter Selection}\label{app:hp_selection}
\begin{figure}
    \centering
    \begin{subfigure}[b]{0.45\textwidth}
        \centering
        \includegraphics[width=\textwidth]{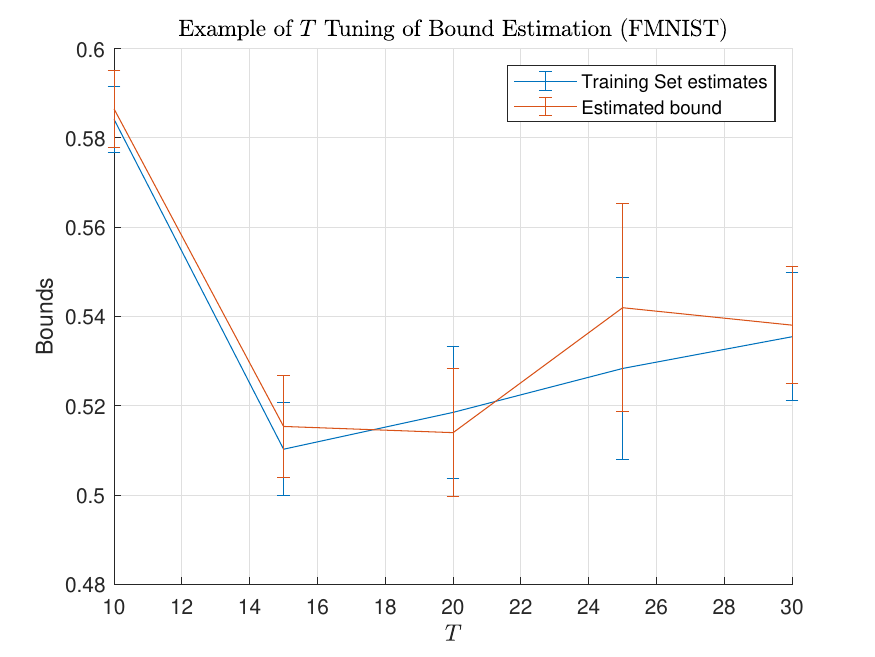}
    \end{subfigure}
        \caption{\textbf{An example of the selection of hyper-parameter $T$}. In this example, The non-adversarial error rate is 0.4, and the trace distance attack strength is 0.1315. The diagram shows the estimates of bounds with different numbers of hyper-spheres that are used to approximate the error region. The blue line indicates the average estimated volume of the $\epsilon$ expansion of the error region for the training set ($S_\textrm{train}$ in Algorithm \ref{alg:loops}), and the orange line indicates the average volume for the testing set ($S_\textrm{test}$ in Algorithm \ref{alg:loops}). In the ideal scenario, we shall find an integer $T$ that minimizes the volume for the testing set as the best choice of $T$.}
    \label{fig:bound_demo}
\end{figure}

During the calculation of the estimated bound, we used different total sphere numbers $T$. As mentioned in Sec.\;\ref{sec:bound}, The bound estimation problem can be formulated as an optimization problem where we minimize the sample number in the $\epsilon$ expansion of the error region parameterized by centers and radii. Thus, the total number of spheres we used to approximate the error region is a hyper-parameter that affects the heuristic solution given by our bound-finding algorithms. In general, for datasets with a more complex data structure, a higher $T$ is required to find the best solution for the algorithm. However, when $T$ is set too high, the generalization of the error region will degrade. For higher $T$, each sphere that constructs the error region contains a smaller number of samples, thus becoming less reliable statistically for estimating the probability where a sample drawn from the underlying data distribution falls in the sphere. This phenomenon of generalization degradation can be observed in the MNIST grid-search result of hyperparameter $T$ in Fig.\;\ref{fig:bound_demo}.

\bibliographystyle{ieeetr}
\bibliography{ref_new_manu}

\end{document}